\documentclass[conference]{IEEEtran}

\IEEEoverridecommandlockouts
\usepackage{cite}
\usepackage{graphicx}
\usepackage{xcolor} 
\usepackage{hyperref}

\usepackage{longtable}
\usepackage{multirow} 
\usepackage{tcolorbox}
\usepackage{enumitem}
\usepackage{arydshln}

\definecolor{blond}{rgb}{0.98, 0.94, 0.75}
\definecolor{lightgreen}{rgb}{0.5, 1.0, 0.83}
\definecolor{pastelyellow}{rgb}{0.99, 0.99, 0.59}
\definecolor{pastelorange}{rgb}{1.0, 0.7, 0.28}
\usepackage{float}
\usepackage{soul}
\usepackage{enumitem}
\usepackage[table,xcdraw]{xcolor}
\usepackage[normalem]{ulem}
\useunder{\uline}{\ul}{}
\usepackage{pifont}

\usepackage{tikz}
\usetikzlibrary{arrows.meta,positioning,fit,calc,shapes.geometric}

\newcommand{\cmark}{\textcolor{green}{\huge \ding{51}}}
\newcommand{\xmark}{\textcolor{orange}{\huge \ding{55}}}

\def\tsc#1{\csdef{#1}{\textsc{\lowercase{#1}}\xspace}}
\tsc{WGM}
\tsc{QE}
\tsc{EP}
\tsc{PMS}
\tsc{BEC}
\tsc{DE}

\usepackage{amsmath,amssymb,amsfonts}
\usepackage{textcomp}
\usepackage{subfigure}
\usepackage{enumerate}
\usepackage[flushleft]{threeparttable}
\usepackage{algorithm}
\usepackage{algorithmic}
\usepackage{float}
\usepackage{diagbox}
\usepackage{booktabs}
\usepackage{adjustbox}
\usepackage{comment}

\usepackage{tabularx}

\makeatletter 
\newcommand{\linebreakand}{%
  \end{@IEEEauthorhalign}
  \hfill\mbox{}\par
  \mbox{}\hfill\begin{@IEEEauthorhalign}
}
\makeatother 

\def\BibTeX{{\rm B\kern-.05em{\sc i\kern-.025em b}\kern-.08em
    T\kern-.1667em\lower.7ex\hbox{E}\kern-.125emX}}
\begin{document}


\title{LLM-ReSum: A Framework for LLM  Reflective Summarization through Self-Evaluation 
\\ {\large Invited Paper}
}


\author{\IEEEauthorblockN{1\textsuperscript{st} Huyen Nguyen}
\IEEEauthorblockA{\textit{Cigna Group} \\
\textit{Evernorth Health Services}\\
Austin, Texas, USA \\
HuyenNguyen5@my.unt.edu}

\and
\IEEEauthorblockN{2\textsuperscript{nd} Haoxuan Zhang}
\IEEEauthorblockA{\textit{dept. of Information Science} \\
\textit{University of North Texas}\\
Denton, Texas, USA \\
haoxuanzhang@my.unt.edu}
\and

\IEEEauthorblockN{3\textsuperscript{rd} Yang Zhang}
\IEEEauthorblockA{\textit{dept. of Data Science} \\
\textit{University of North Texas}\\
Denton, Texas, USA \\
yang.zhang@unt.edu}
\and

\linebreakand 

\IEEEauthorblockN{4\textsuperscript{th} Haihua Chen}
\IEEEauthorblockA{\textit{dept. of Data Science} \\
\textit{University of North Texas}\\
Denton, Texas, USA \\
haihua.chen@unt.edu}
\and

\IEEEauthorblockN{5\textsuperscript{th} Junhua Ding\IEEEauthorrefmark{1}\thanks{This work was
supported in part by the NSF under Grants \#2225229, \#2601493, and \#2231519. Corresponding author: Junhua Ding (Email: junhua.ding@unt.edu)}}
\IEEEauthorblockA{\textit{dept. of Data Science} \\
\textit{University of North Texas}\\
Denton, Texas, USA \\
junhua.ding@unt.edu}
}

\maketitle


\thispagestyle{plain}
\pagestyle{plain}

\begin{abstract}
Reliable evaluation of large language model (LLM)–generated summaries remains an open challenge, particularly across heterogeneous domains and document lengths. We conduct a comprehensive meta-evaluation of 14 automatic summarization metrics and LLM-based evaluators across seven datasets spanning five domains, covering documents from short news articles to long scientific, governmental, and legal texts (2K–27K words) with over 1,500 human-annotated summaries. Our results show that traditional lexical overlap metrics (e.g., ROUGE, BLEU) exhibit weak or negative correlation with human judgments, while task-specific neural metrics and LLM-based evaluators achieve substantially higher alignment, especially for linguistic quality assessment. Leveraging these findings, we propose \textbf{LLM-ReSum}, a self-reflective summarization framework that integrates LLM-based evaluation and generation in a closed feedback loop without model finetuning. Across three domains, LLM-ReSum improves low-quality summaries by up to 33\% in factual accuracy and 39\% in coverage, with human evaluators preferring refined summaries in 89\% of cases. We additionally introduce \textbf{PatentSumEval}, a new human-annotated benchmark for legal document summarization comprising 180 expert-evaluated summaries. All code and datasets will be released in GitHub.
\end{abstract}


\section{Introduction}
\label{intro}
Text summarization has emerged as an indispensable technology for enabling efficient information consumption across diverse domains, including news media, governmental documentation, scientific literature, social platforms, and legal corpora~\cite{10.1145/3731445, huang2025survey}. Modern summarization systems, increasingly powered by large language models (LLMs), are now deployed at scale in search engines, recommendation systems, and knowledge extraction pipelines \cite{10.1145/3731445, zhang2022comprehensive, huang2025survey}. The proliferation of these systems has intensified the demand for robust evaluation methodologies that align with human quality judgments \cite{gao2023human, 10.1162/tacl_a_00373, ding2023quality}.

Conventional automatic evaluation metrics, such as n-gram overlap measures (ROUGE, BLEU), embedding-based similarity scores (BERTScore), and task-specific NLP models, have dominated summarization assessment for decades. Traditional approaches have employed topic-based vector space models and semantic measures for evaluation \cite{belwal2021text}, while extractive methods have focused on candidate sentence selection strategies \cite{mutlu2020candidate}. Nevertheless, accumulating evidence reveals systematic failures of these metrics to capture dimensions of quality prioritized by human evaluators, particularly factual consistency, semantic coverage, and linguistic fluency \cite{10.1162/tacl_a_00373, zhangbertscore}. Lexical overlap metrics such as ROUGE and BLEU exhibit fundamental limitations in semantic understanding, penalizing paraphrases and rewarding superficial word matches \cite{10.1162/coli_a_00322}. These shortcomings are especially pronounced for abstractive summarization, where neural models generate novel phrasings rather than extractive fragments \cite{belwal2021text}. Notably, while human evaluators substantially prefer summaries from prompt-based LLMs such as GPT-3, these outputs receive significantly lower ROUGE scores than earlier fine-tuned models, exposing a critical misalignment between automatic metrics and human preferences \cite{goyal2023news, ding2024evaluation, nguyen2024comparative}. This discrepancy raises fundamental questions about metric validity across heterogeneous domains, document lengths, and summarization paradigms.

Concurrently, recent advancements in LLMs have enabled a paradigm shift: leveraging LLMs as evaluation agents. Given their capacity to process nuanced natural language instructions and assess multiple quality dimensions simultaneously, LLMs present a compelling alternative to handcrafted metrics. Empirical investigations have demonstrated that ChatGPT and similar models, when employed for Likert-scale rating or pairwise comparison, can exceed conventional metrics in correlation with human judgments \cite{gao2023human, luo2023chatgpt, liu-etal-2023-g}. Recent work has also explored LLM feedback mechanisms for improving text generation quality \cite{song2025causal}. However, LLM-based evaluation exhibits significant sensitivity to prompt engineering and domain shift \cite{gao2023human}, with existing research predominantly concentrated on news corpora. Critical gaps persist regarding cross-domain robustness, quantitative benchmarking against traditional metrics, and reliability across diverse document types including domain-specific applications such as biomedical literature, patent documents, and legal cases \cite{huang2025survey, ding2023quality, akter2025comprehensive}.

Beyond their role as evaluators, LLMs offer an opportunity to establish a closed-loop framework integrating evaluation and generation. Iterative refinement driven by evaluation feedback could enable LLMs to self-correct outputs toward greater human alignment without resource-intensive finetuning \cite{song2025causal}. While reinforcement learning from human feedback has proven effective \cite{NEURIPS2020_1f89885d}, such approaches demand substantial computational infrastructure and curated preference data. Prompt-based self-improvement presents a more scalable alternative, yet poses unresolved challenges in prompt design, feedback representation, convergence criteria, and error amplification mitigation.

The primary objective of this study is to systematically examine the reliability of existing summarization evaluation methods in the context of large language model (LLM)–generated summaries and to investigate whether evaluation signals can be effectively leveraged to improve summarization quality. Specifically, this research aims (1) to assess the extent to which widely used automatic evaluation metrics align with human judgments across diverse domains and document lengths; (2) to evaluate the effectiveness and robustness of LLM-based evaluation approaches, including single-agent and multi-agent configurations, as alternatives to conventional metrics; and (3) to develop and empirically validate a self-reflective summarization framework that integrates evaluation and generation to enable iterative quality improvement without model finetuning. To address these objectives, we formulate three research questions (RQs):

\begin{itemize}
    \item \textbf{RQ1:} To what extent do traditional automatic evaluation metrics correlate with human judgments across diverse domains and document lengths?
    \item \textbf{RQ2:} How reliably can LLMs function as evaluation metrics for summarization quality assessment compared to conventional approaches?
    \item \textbf{RQ3:} Can a human-aligned LLM evaluation framework enable iterative self-refinement to improve summarization quality?
\end{itemize}

To answer RQ1, we conduct a systematic meta-evaluation of 14 widely adopted automatic metrics across seven datasets spanning five domains with varying input lengths, quantifying their alignment with human judgments on multiple quality dimensions. For RQ2, we investigate single-agent and multi-agent LLM evaluation architectures, benchmarking their performance against conventional metrics and analyzing domain-specific strengths and limitations. Addressing RQ3, we introduce \textbf{LLM-ReSum}, a novel reflective summarization framework wherein LLMs iteratively refine outputs based on self-generated evaluation feedback aligned with human quality criteria. Validation through both automatic metrics and human evaluation demonstrates substantial quality improvements, particularly in factual accuracy and semantic coverage.

Our contributions are fourfold:
\begin{itemize}
\item We conduct a comprehensive meta-evaluation of widely-used automatic metrics and LLM-based evaluation across seven datasets spanning five domains, revealing systematic failures of traditional lexical overlap metrics for evaluating LLM-generated summaries and establishing empirical reliability benchmarks across diverse contexts.

\item We systematically compare single-agent and multi-agent LLM evaluation architectures, demonstrating that multi-agent frameworks significantly outperform conventional metrics on linguistic quality dimensions, providing clear guidance for selecting appropriate evaluation strategies in practical deployment.

\item We propose \textbf{LLM-ReSum}, a novel self-reflective summarization framework integrating LLM evaluation and generation in a closed feedback loop, achieving substantial gains of up to 33\% in accuracy and 39\% in coverage with 89\% human preference rate, without requiring model finetuning or reinforcement learning.

\item We introduce \textbf{PatentSumEval}, a human-annotated benchmark for legal patent document summarization evaluation, and release reproducible meta-evaluation code to facilitate future research.
\end{itemize}

\section{Related Work}

\subsection{Evaluation Methods for Summarization}

Summarization evaluation approaches are broadly categorized as reference-based or reference-free \cite{10.1162/tacl_a_00373}. Reference-based methods measure similarity against human-authored summaries, while reference-free approaches assess quality directly from source documents. Given the prohibitive cost of reference creation, particularly in specialized domains \cite{huang2025survey}, we focus on reference-free evaluation.

Automatic evaluation metrics dominate current practice. N-gram overlap metrics (ROUGE \cite{lin2004rouge}, BLEU \cite{10.3115/1073083.1073135}) quantify lexical similarity through surface-form matching. Embedding-based metrics (BERTScore \cite{zhangbertscore}, MoverScore \cite{zhao-etal-2019-moverscore}) leverage contextualized representations to capture semantic correspondence. Task-specific metrics operationalize quality through auxiliary NLP tasks: SummaC \cite{laban2022summac} employs natural language inference for consistency verification, while QuestEval \cite{scialom-etal-2021-questeval} utilizes question-answering frameworks. These metrics exhibit systematic deficiencies. Empirical studies reveal weak correlations between automated scores and human quality judgments \cite{10.1162/tacl_a_00373, sun-etal-2022-bertscore}, particularly for lexical metrics that penalize semantically valid paraphrases \cite{10.1162/coli_a_00322}. 

Given these limitations, researchers have conducted meta-evaluations to assess which metrics most reliably correlate with human judgments. However, existing reliability assessments suffer from a narrow scope. Prior meta-evaluations typically examine one or two datasets within homogeneous domains \cite{bhandari-etal-2020-evaluating, 10.1162/tacl_a_00373, koh-etal-2022-far}, constraining generalizability of their findings. Consequently, cross-domain robustness and performance across varying document characteristics remain uncharacterized.

\subsection{LLMs in Summarization: Evaluation and Improvement}

The emergence of instruction-tuned LLMs has introduced new paradigms for both evaluation and quality improvement in summarization. Instruction-following capabilities enable direct quality assessment through natural language specifications of evaluation criteria. ChatGPT demonstrates human-comparable performance on Likert scoring and pairwise ranking tasks, surpassing conventional metrics \cite{gao2023human}. Zero-shot consistency detection \cite{luo2023chatgpt} and chain-of-thought evaluation frameworks \cite{liu-etal-2023-g} achieve superior human alignment on established benchmarks. Domain-specific applications, including biomedical summarization \cite{huang2025survey}, further validate LLM evaluator capabilities. Despite these advances, critical limitations persist: evaluation quality exhibits high prompt sensitivity \cite{mahmoudi-2023-exploring} and domain dependence \cite{gao2023human}, research concentrates predominantly on news corpora while specialized domains remain underexplored, and single-agent versus multi-agent architectures lack systematic comparison across diverse contexts.

Beyond evaluation, parallel research investigates LLM-driven quality improvement through two paradigms. Reinforcement learning methods optimize generation through feedback signals: early approaches incorporated ROUGE rewards \cite{paulus2018a, 10.5555/3304222.3304389}, while recent work employs human preference models \cite{stanczak2025societal} or LLM-synthesized feedback \cite{song-etal-2025-learning}. Despite effectiveness, these methods demand substantial computational resources and curated training data. Alternatively, prompt-based adaptation enables task-specific behavior through structured instructions without parameter modification \cite{brown2020lan}. Recent investigations of feedback-driven text refinement \cite{song2025causal} demonstrate viability for iterative improvement, though applications to summarization remain limited.

Existing work treats evaluation and improvement as independent processes. While reinforcement learning approaches integrate feedback during training \cite{stanczak2025societal, song-etal-2025-learning}, and prompt-based methods enable iterative refinement \cite{song2025causal}, no prior framework integrates LLM evaluation and generation in a closed feedback loop for iterative summarization quality enhancement. Specifically, existing evaluation studies focus on metric reliability \cite{gao2023human, luo2023chatgpt, liu-etal-2023-g} without exploring how evaluation outcomes can guide revision, and improvement methods rely on external feedback sources \cite{stanczak2025societal} or general-purpose refinement strategies \cite{song2025causal} rather than summary-specific evaluation criteria. Key questions remain unaddressed: whether evaluation feedback can be translated into actionable revision guidance, which quality dimensions respond to iterative refinement, and whether self-correction converges toward improved quality. We address these gaps by proposing \textbf{LLM-ReSum}, a self-reflective architecture wherein LLMs evaluate outputs against human-aligned criteria and leverage generated feedback for iterative refinement, achieving quality improvements without model finetuning.

\section{Methodology}

\subsection{Framework Overview}

\begin{figure*}[ht]
    \centering
    \includegraphics[width=0.8\linewidth]{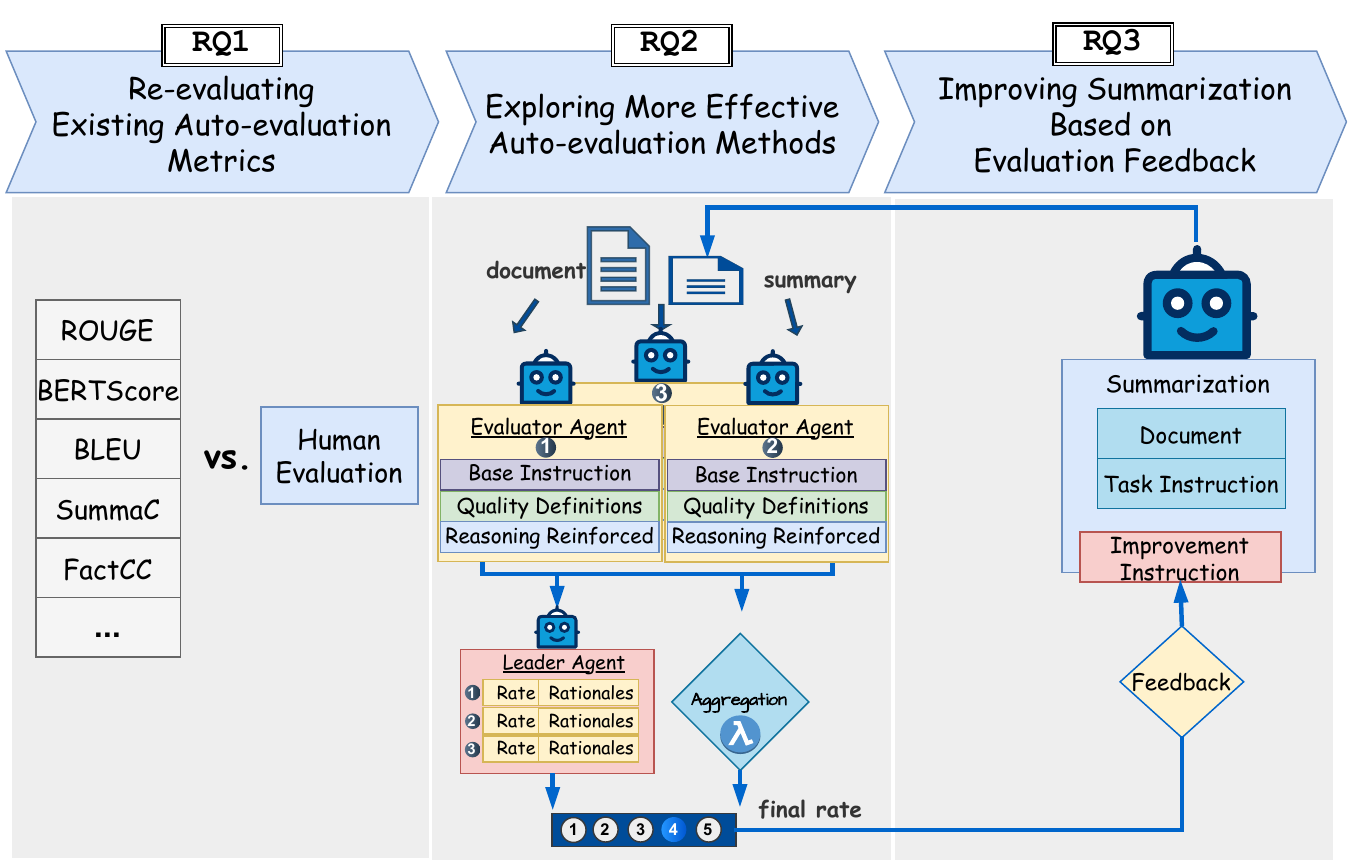}
    \caption{Overview of our three-stage research framework: meta-evaluation of automatic metrics (RQ1), multi-agent LLM evaluation (RQ2), and iterative self-reflective summarization (RQ3).}
    \label{fig:paper-workflow}
\end{figure*}

This study adopts a three-stage methodological framework to systematically address the research questions posed in Section~\ref{intro}. To answer RQ1, we employ meta-evaluation methodology to assess the reliability of automatic evaluation metrics by measuring their correlation with human judgments across diverse contexts. For RQ2, we design single-agent and multi-agent LLM evaluation frameworks that leverage instruction-following capabilities and chain-of-thought reasoning to simulate human evaluation behavior. To address RQ3, we propose \textbf{LLM-ReSum}, a self-reflective summarization framework that integrates evaluation and generation in a closed feedback loop to enable iterative quality improvement without model finetuning. Figure~\ref{fig:paper-workflow} illustrates the complete research workflow.

\subsection{Task Formulation}

We formalize the reference-free summarization evaluation and iterative refinement tasks as follows. Let \(D\) denote a source document and \(S\) denote a candidate summary. The evaluation task computes quality scores across \(K\) dimensions \(\mathcal{Q} = \{q_1, q_2, \ldots, q_K\}\):

\begin{equation}
\centering
\mathbf{q}(S, D) = [q_1(S, D), q_2(S, D), \ldots, q_K(S, D)]
\label{eq:eval-function}
\end{equation}

where each \(q_i(S, D) \in [1, L]\) represents a score on a Likert scale with \(L\) levels.

For iterative refinement, starting with an initial summary \(S^{(0)} = f_{\text{sum}}(D)\), we iteratively improve quality through evaluation-guided feedback:

\begin{equation}
\centering
S^{(t+1)} = f_{\text{refine}}(S^{(t)}, D, F^{(t)})
\label{eq:iterative-refine}
\end{equation}

where \(F^{(t)} = g_{\text{feedback}}(\mathbf{q}^{(t)}, \mathbf{r}^{(t)})\) represents structured feedback constructed from evaluation scores \(\mathbf{q}^{(t)}\) and rationales \(\mathbf{r}^{(t)}\) at iteration \(t\). The process terminates when \(\min_i q_i^{(t)} \geq \tau\) (quality threshold satisfied) or \(t = T_{\text{max}}\) (maximum iterations reached).

\subsection{LLM-Based Evaluation Framework}

\subsubsection{Direct Evaluation Paradigm}

We adopt the direct evaluation paradigm wherein evaluators assess generated summaries according to predefined quality criteria without relying on reference summaries. This approach offers several advantages: it avoids assuming reference summaries represent perfect quality, enables evaluation along dimensions not captured by reference similarity, and provides flexibility to customize evaluation criteria for domain-specific applications \cite{gao2023human}.

Given that LLMs demonstrate exceptional instruction-following capabilities and can process complex natural language specifications of evaluation criteria \cite{ouyang2022training}, we hypothesize they can emulate human evaluation behavior when provided with appropriately structured prompts \cite{liu-etal-2023-g, kocmi-federmann-2023-large}. We formalize LLM-based evaluation as \(\mathbf{q}_{\text{LLM}}(S, D) = f_{\text{LLM}}(P_{\text{eval}}, S, D)\), where \(P_{\text{eval}}\) encodes evaluation instructions and quality dimension definitions.

\subsubsection{Multi-Agent Evaluation Framework}

Inspired by human evaluation practices that aggregate judgments from multiple annotators to mitigate individual bias \cite{10.1145/1233912.1233913}, we propose a multi-agent evaluation framework employing heterogeneous LLMs as independent evaluation agents. Each agent independently evaluates summaries using the Reasoning Reinforced prompt strategy, producing dimension-specific scores and rationales.

When agents produce divergent assessments, we explore three aggregation mechanisms:

\textbf{Averaging} computes the arithmetic mean of scores across agents for each dimension, treating all agents equally and producing continuous scores.

\textbf{Majority Voting} identifies the most frequent score among agents for each dimension. In cases of three-way disagreement, the median score is selected. This method enforces discrete scoring and emphasizes consensus \cite{wang-etal}.

\textbf{Leader-Based Aggregation} introduces an additional LLM serving as a meta-evaluator that receives independent evaluations and rationales, analyzes points of agreement and disagreement, weighs the strength of different rationales, and produces a final assessment \cite{du2023improving}. This approach leverages meta-reasoning capabilities to resolve conflicts intelligently.

\subsection{LLM-ReSum: Self-Reflective Summarization Framework}

Building upon the LLM evaluation framework, we propose \textbf{LLM-ReSum}—a novel architecture that integrates evaluation and generation in a closed feedback loop for iterative quality improvement. The framework addresses a fundamental limitation of conventional summarization: the disconnect between training objectives and evaluation criteria \cite{song-etal-2025-learning}. By incorporating evaluation directly into the generation process through prompt-based feedback, LLM-ReSum enables zero-shot quality improvement without requiring supervised finetuning or reinforcement learning \cite{brown2020lan, welleck2023generating}. The framework operationalizes two key hypotheses. First, explicit evaluation feedback aligned with human quality criteria can guide LLMs toward self-correction \cite{bai2022constitutional, NEURIPS2023_91edff07}. Second, iterative refinement through evaluation-generation cycles can progressively improve output quality when properly constrained to prevent degradation \cite{welleck2023generating}.

\subsubsection{Iterative Refinement Process}

\textbf{LLM-ReSum} executes a four-stage iterative cycle (Algorithm~\ref{alg:llm-resum}): (1) generate initial summary using standard summarization instructions (Prompt 1), (2) evaluate the summary across multiple quality dimensions producing scores and rationales using our reasoning-reinforced evaluation prompt ( Prompt 2) \cite{liu-etal-2023-g}, (3) construct actionable feedback for dimensions scoring below threshold \(\tau\), and (4) generate refined summary conditioned on the source document and evaluation feedback using the refinement prompt (Prompt 3) \cite{NEURIPS2023_91edff07}. The cycle continues until convergence criteria are satisfied or maximum iterations \(T_{\text{max}}\) are exhausted.

\begin{algorithm}[t]
\centering
\caption{LLM-ReSum: Self-Reflective Summarization}
\label{alg:llm-resum}
\begin{algorithmic}[1]
\REQUIRE Source document \(D\), Summarization function \(f_{\text{sum}}\), Evaluation function \(f_{\text{eval}}\), Refinement function \(f_{\text{refine}}\), Quality threshold \(\tau\), Maximum iterations \(T_{\text{max}}\)
\ENSURE Enhanced summary \(S^*\)
\STATE \(S^{(0)} \gets f_{\text{sum}}(D)\) \COMMENT{Generate initial summary}
\STATE \(t \gets 0\)
\WHILE{\(t < T_{\text{max}}\)}
    \STATE \(\mathbf{q}^{(t)}, \mathbf{r}^{(t)} \gets f_{\text{eval}}(D, S^{(t)})\) \COMMENT{Evaluate current summary}
    \IF{\(\min_i q_i^{(t)} \geq \tau\)}
        \RETURN \(S^{(t)}\) \COMMENT{Quality threshold satisfied}
    \ENDIF
    \STATE \(\mathcal{I} \gets \{i \mid q_i^{(t)} < \tau\}\) \COMMENT{Identify deficient dimensions}
    \STATE \(F^{(t)} \gets g_{\text{feedback}}(\{q_i^{(t)}, r_i^{(t)}\}_{i \in \mathcal{I}})\) \COMMENT{Construct feedback}
    \STATE \(S^{(t+1)} \gets f_{\text{refine}}(D, F^{(t)})\) \COMMENT{Generate refined summary}
    \STATE \(t \gets t + 1\)
\ENDWHILE
\RETURN \(\arg\max_{t'} \min_i q_i^{(t')}\) \COMMENT{Return best summary}
\end{algorithmic}
\end{algorithm}

{\small
\begin{tcolorbox}[
    colback=blue!5, 
    colframe=blue!50!black, 
    boxrule=1pt, 
    rounded corners, 
    title=\textbf{Prompt 1: Initial Summarization}, 
    fonttitle=\bfseries,
    label={prompt:initial}
]

\textbf{System Instruction}\\
You are an expert summarization system specialized in creating concise, accurate, and comprehensive summaries.

\vspace{4pt}
\textbf{Task Description} \\
Generate a high-quality summary of the document provided below. The summary should capture the essential information while maintaining clarity and accuracy.

\vspace{4pt}
\textbf{Requirements}

\begin{itemize}[leftmargin=1.5em, itemsep=2pt, parsep=0pt, topsep=2pt]
    \item Include all key points and important information
    \item Maintain factual accuracy with the source document
    \item Use clear and concise language
    \item Ensure the summary is reader-friendly
\end{itemize}

\vspace{4pt}
\textbf{Source Document} \\
\textit{\{original\_document\}}

\vspace{4pt}
\textbf{Output Format} \\
Provide only the summary without additional commentary.

\end{tcolorbox}
}

{\small
\begin{tcolorbox}[
    colback=orange!5, 
    colframe=orange!50!black, 
    boxrule=1pt, 
    rounded corners, 
    title=\textbf{Prompt 2: Summary Evaluation (Reasoning-Reinforced)}, 
    fonttitle=\bfseries,
    label={prompt:evaluation}
]

\textbf{System Instruction}\\
You are an expert evaluator specializing in assessing summarization quality.

\vspace{4pt}
\textbf{Task Description} \\
Evaluate the quality of the summary below against the original document across multiple quality dimensions.

\vspace{4pt}
\textbf{Quality Dimensions}

\begin{itemize}[leftmargin=1.5em, itemsep=2pt, parsep=0pt, topsep=2pt]
    \item \textbf{Clarity}: Is the summary reader-friendly? Does it express ideas clearly?
    \item \textbf{Accuracy}: Does the summary contain the same information as the original document?
    \item \textbf{Coverage}: How well does the summary cover the important information in the original document?
    \item \textbf{Overall quality}: How good is the summary overall at representing the original document?
\end{itemize}

\vspace{4pt}
\textbf{Evaluation Instructions}

On a scale of 1-5, rate the summary on each dimension above. \textbf{Provide one to two short sentences to explain your rating for each dimension}, citing specific evidence from the summary.

\vspace{4pt}
\textbf{Source Document} \\
\textit{\{original\_document\}}

\vspace{4pt}
\textbf{Summary to Evaluate} \\
\textit{\{summary\_to\_evaluate\}}

\vspace{4pt}
\textbf{Output Format} \\
The output must follow the following Python dictionary format:\\
\texttt{\{'clarity': clarity\_score, 'accuracy': accuracy\_score, 'coverage': coverage\_score, 'overall': overall\_score, 'explanation': \{'clarity': explanation, 'accuracy': explanation, 'coverage': explanation, 'overall': explanation\}\}}

\end{tcolorbox}
}

{\small
\begin{tcolorbox}[
    colback=green!5, 
    colframe=green!50!black, 
    boxrule=1pt, 
    rounded corners, 
    title=\textbf{Prompt 3: Summary Refinement}, 
    fonttitle=\bfseries,
    label={prompt:refinement}
]

\textbf{System Instruction}\\
You are an expert summarization system. Your task is to generate an improved summary by addressing quality issues identified in the previous evaluation.

\vspace{4pt}
\textbf{Task Description} \\
Revise the summary below to resolve all flagged deficiencies while maintaining factual accuracy and completeness.

\vspace{4pt}
\textbf{Quality Dimensions}

\begin{itemize}[leftmargin=1.5em, itemsep=2pt, parsep=0pt, topsep=2pt]
    \item \textbf{Clarity}: Is the summary reader-friendly? Does it express ideas clearly?
    \item \textbf{Accuracy}: Does the summary contain the same information as the original document?
    \item \textbf{Coverage}: How well does the summary cover the important information in the original document?
    \item \textbf{Overall quality}: How good is the summary overall at representing the original document?
\end{itemize}

\vspace{4pt}
\textbf{Evaluation Feedback from Previous Summary}

The previous summary received the following assessment (scale 1-5):

\begin{itemize}[leftmargin=1.5em, itemsep=2pt, parsep=0pt, topsep=2pt]
    \item \textbf{Clarity} [Score: \textit{\{score\}}]: \textit{\{clarity\_explanation\}}
    \item \textbf{Accuracy} [Score: \textit{\{score\}}]: \textit{\{accuracy\_explanation\}}
    \item \textbf{Coverage} [Score: \textit{\{score\}}]: \textit{\{coverage\_explanation\}}
    \item \textbf{Overall quality} [Score: \textit{\{score\}}]: \textit{\{overall\_explanation\}}
\end{itemize}

\vspace{4pt}
\textbf{Instructions}

\begin{itemize}[leftmargin=1.5em, itemsep=0pt, parsep=0pt, topsep=2pt]
    \item Address each issue identified in the evaluation feedback above
    \item Ensure the revised summary scores at least 4/5 on all dimensions
    \item Verify all information against the source document below
    \item Maintain conciseness while improving quality
\end{itemize}

\vspace{4pt}
\textbf{Source Document} \\
\textit{\{original\_document\}}

\vspace{4pt}
\textbf{Previous Summary} \\
\textit{\{summary\_to\_evaluate\}}

\vspace{4pt}
\textbf{Output Format} \\
Provide only the improved summary without additional commentary.

\end{tcolorbox}
}

\subsubsection{Feedback Construction and Convergence}

A critical innovation is the transformation of evaluation scores into actionable improvement guidance. For each dimension scoring below threshold \(\tau\), we extract the evaluation rationale and format it as structured natural language feedback. This approach serves dual purposes: it explicitly identifies deficient quality aspects and provides concrete error examples for correction. Unlike generic instructions, dimension-specific feedback with detailed rationales enables targeted revisions aligned with human quality expectations \cite{ouyang2022training}. 

The refinement prompt (Prompt 3) comprises three components: (1) base instruction directing quality improvement with explicit dimension definitions, (2) evaluation feedback presenting dimension-specific scores and explanatory rationales from the evaluation phase, and (3) the source document for fact verification.

We establish convergence criteria to balance quality improvement against iterative degradation risk \cite{welleck2023generating}. The quality threshold is set to \(\tau = 4\) on a 5-point Likert scale, where scores of 4 (``Good'') and 5 (``Excellent'') indicate acceptable quality. Refinement is triggered when any dimension scores below \(\tau\), ensuring comprehensive quality assurance across all evaluated aspects. We impose maximum iterations \(T_{\text{max}} = 3\) to prevent excessive refinement that may introduce semantic drift, over-compression, or hallucinations \cite{maynez-etal}. If convergence is not achieved within the iteration limit, the system returns the summary with the highest minimum quality score across all dimensions, ensuring the best available output is selected.

\section{Datasets}

\begin{figure}
    \centering
    \includegraphics[width=\linewidth]{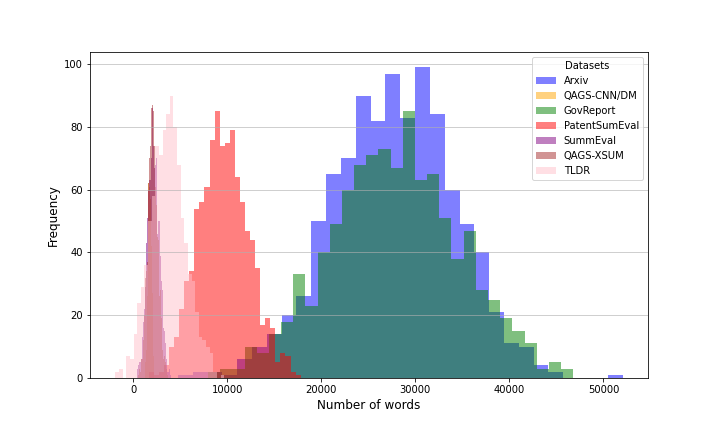}
    \caption{Source document length distributions}
    \label{fig:length-source}
\end{figure}

\begin{figure}
    \centering
    \includegraphics[width=\linewidth]{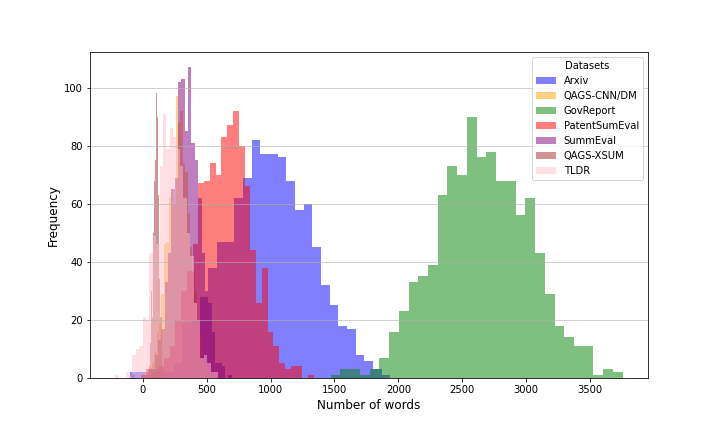}
    \caption{Model-generated summary length distributions}
    \label{fig:length-summary}
\end{figure}

We conduct our meta-evaluation across seven datasets spanning five diverse domains to comprehensively assess metric reliability across different text types and lengths. Table~\ref{tab:ds-human-eval} presents an overview of these datasets. Six datasets are drawn from existing literature, while we introduce PatentSumEval as a novel benchmark for legal domain summarization evaluation. The datasets exhibit substantial variation in input document length, ranging from short news articles (average 2,084 words in QAGS-XSUM) to lengthy research papers and government reports (average 27,631 words in GovReport), as illustrated in Figure~\ref{fig:length-source} and Figure~\ref{fig:length-summary}, enabling robust assessment of metric performance across diverse summarization contexts.

\begin{table*}[t]
\centering
\caption{Overview of meta-evaluation datasets with human annotations. The datasets span five domains with varying document lengths and evaluation protocols.}
\label{tab:ds-human-eval}
\resizebox{1\textwidth}{!}{%
\begin{tabular}{l|c|c|l|l|cc|cccc}
\hline
  \multirow{2}{*}{\textbf{Dataset}} &
  \multirow{2}{*}{\textbf{\# Models}} &
  \multirow{2}{*}{\textbf{\# Docs}} &
  \multirow{2}{*}{\textbf{Domain}} &
  \multirow{2}{*}{\textbf{Human Eval.}} &
  \multicolumn{2}{c|}{\textbf{Length (words)}} &
  \multicolumn{4}{c}{\textbf{Evaluation Dimensions}} \\ \cline{6-11}    
  &   &   &   &   &
   \multicolumn{1}{c}{\textbf{Source}} &
 \multicolumn{1}{c|}{\textbf{Summary}} &
  \multicolumn{1}{c|}{\textbf{Ling.}} &
  \multicolumn{1}{c|}{\textbf{Acc./Cons.}} &
  \multicolumn{1}{c|}{\textbf{Cov./Rel.}} &
  \textbf{Overall} \\ \hline
  SummEval~\cite{10.1162/tacl_a_00373} &
  23 &
  100 &
  News &
  Likert (1-5) &
  2,160 (601) & 
  343 (105) &
  \multicolumn{1}{c|}{\cmark} &
  \multicolumn{1}{c|}{\cmark} &
  \multicolumn{1}{c|}{\cmark} &
  \xmark \\ \hline
  Arxiv~\cite{koh-etal-2022-far} &
  12 &
  17 &
  Scientific &
  Binary$^{\dagger}$ &
  27,193 (6,558) &
  972 (342) &
  \multicolumn{1}{c|}{\xmark} &
  \multicolumn{1}{c|}{\cmark} &
  \multicolumn{1}{c|}{\cmark} &
  \xmark \\ \hline
  GovReport~\cite{koh-etal-2022-far} &
  12 &
  17 &
  Government &
  Binary$^{\dagger}$ &
  27,631 (6,943) &	
  2,658 (363) &
  \multicolumn{1}{c|}{\xmark} &
  \multicolumn{1}{c|}{\cmark} &
  \multicolumn{1}{c|}{\cmark} &
  \xmark \\ \hline
  TLDR~\cite{10.5555/3495724.3495977} &
  8 &
  788 &
  Social media &
  Likert (1-7) &
  3,756 (1,866) &
  227 (123) &
  \multicolumn{1}{c|}{\cmark} &
  \multicolumn{1}{c|}{\cmark} &
  \multicolumn{1}{c|}{\cmark} &
  \cmark \\ \hline
  QAGS-XSUM~\cite{wang-etal-2020-asking} &
  1 &
  239 &
  News &
  Binary$^{\ddagger}$ &
  2,084 (514) &
  105 (22) &
  \multicolumn{1}{c|}{\xmark} &
  \multicolumn{1}{c|}{\cmark} &
  \multicolumn{1}{c|}{\xmark} &
  \xmark \\ \hline
  QAGS-CNN/DM~\cite{wang-etal-2020-asking} &
  1 &
  235 &
  News &
  Binary$^{\ddagger}$ &
  1,791 (192) &
  284 (83) &
  \multicolumn{1}{c|}{\xmark} &
  \multicolumn{1}{c|}{\cmark} &
  \multicolumn{1}{c|}{\xmark} &
  \xmark \\ \hline
  PatentSumEval$^{*}$ &
  6 &
  30 &
  Legal &
  Likert (1-5) &
  9,754 (2,727) &
  647 (209) &
  \multicolumn{1}{c|}{\cmark} &
  \multicolumn{1}{c|}{\cmark} &
  \multicolumn{1}{c|}{\cmark} &
  \cmark \\ \hline
\end{tabular}%
}
\vspace{0.1cm}
\begin{flushleft}
\scriptsize
\textit{Note:} $^{*}$Our proposed dataset. $^{\dagger}$Sentence-level binary evaluations averaged to summary-level scores. $^{\ddagger}$Sentence-level binary consistency judgments aggregated via averaging. Ling. = Linguistic quality (coherence, fluency, clarity); Acc./Cons. = Accuracy/Consistency; Cov./Rel. = Coverage/Relevance. Length values represent the mean (standard deviation) in word count. \cmark indicates dimension evaluated; \xmark indicates not evaluated.
\end{flushleft}
\end{table*}

\subsection{Existing Datasets}

\textbf{SummEval:} Contains 100 CNN/DailyMail news articles with summaries from 23 models (16 extractive, 7 abstractive). Five Amazon Mechanical Turk workers and three experts rated summaries on a 5-point Likert scale across coherence, consistency, fluency, and relevance~\cite{10.1162/tacl_a_00373}.

\textbf{Arxiv:} Comprises scientific paper summaries from 12 long-document models including TDT~\cite{pang-etal-2023-long} and DYLE~\cite{mao-etal-2022-dyle}, applied to documents sampled from the arXivPubMed corpus. Three experts provided sentence-level binary judgments for factual consistency and relevance, aggregated to summary-level scores~\cite{koh-etal-2022-far}.

\textbf{GovReport:} Contains government report summaries from the same 12 models as Arxiv, applied to documents from the GovReport dataset. Three experts evaluated summaries using sentence-level binary judgments for factual consistency and relevance, averaged to produce summary-level evaluations~\cite{koh-etal-2022-far}.

\textbf{TLDR:} Consists of 788 Reddit posts with summaries from 8 reinforcement learning model variants. Each summary received ten evaluations on a 7-point Likert scale across coherence, accuracy, coverage, and overall quality~\cite{10.5555/3495724.3495977}.

\textbf{QAGS-XSUM:} Contains 239 news article summaries from the XSUM dataset generated by the BART model. Three Amazon Mechanical Turk workers evaluated factual consistency through sentence-level binary judgments, aggregated via averaging~\cite{wang-etal-2020-asking}.

\textbf{QAGS-CNN/DM:} Comprises 235 news article summaries from the CNN/DailyMail dataset generated by the Bottom-Up model. Three Amazon Mechanical Turk workers assessed factual consistency using sentence-level binary judgments, averaged to summary-level scores~\cite{wang-etal-2020-asking}.

\subsection{PatentSumEval Benchmark (Ours)}

\subsubsection{Dataset Construction}

Since no publicly available datasets exist for evaluating legal document summarization with human annotations, we constructed PatentSumEval to assess metric performance on domain-specific technical documents. The benchmark comprises 180 summaries generated by six models across 30 patent documents related to communication and streaming technologies, collected from Google Patents\footnote{\url{https://patents.google.com}}.

\textbf{Input Selection.} Patent documents contain extensive technical descriptions, but the most critical content includes the abstract and claims. The abstract provides an invention overview, while claims detail specific novel aspects. Unlike existing datasets such as BIGPATENT~\cite{sharma-etal-2019-bigpatent} that use only the abstract as reference, we concatenate both abstract and claims as model input to enable generation of summaries that comprehensively cover both the invention's scope and novelty.

\textbf{Model Selection.} We generated summaries using six diverse models representing different architectural paradigms: HUPD-T5-base (domain-specific patent model), XLNet, BART, and Pegasus (general-purpose pretrained models), LongT5 (long-context specialist), and GPT-3.5-turbo (instruction-tuned LLM). Each model generated one summary per document, yielding 180 summaries total.

\subsubsection{Human Evaluation Protocol}

\textbf{Annotator Recruitment.} Given the technical complexity of communication and engineering patents, we recruited three Master's students in computer science and engineering fields with sufficient domain expertise to assess summary quality. Each annotator independently evaluated all 180 summaries. To ensure annotation reliability, we embedded quality control questions throughout the evaluation to identify and filter inconsistent responses.

\textbf{Evaluation Platform.} We conducted evaluations using the APPEN platform\footnote{\url{https://client.appen.com}}, which provides structured interfaces for Likert-scale assessments and facilitates quality control. The complete evaluation form is publicly available\footnote{\href{https://account.appen.com/channels/cf_internal/jobs/2377354/work?secret=2g7C\%2FLK\%2BeVn\%2FDg\%2FdFFjyjXnTJLz2LC7c8vBLPNOhG6zo}{APPEN evaluation form}}.

\textbf{Quality Dimensions.} Annotators rated each summary on a 5-point Likert scale (1=Poor, 2=Fair, 3=Adequate, 4=Good, 5=Excellent) across four dimensions:
\begin{itemize}[leftmargin=1.5em, itemsep=2pt, parsep=0pt, topsep=2pt]
    \item \textit{Clarity}: Whether the summary expresses ideas clearly without ambiguity or readability issues.
    \item \textit{Accuracy}: Whether all information in the summary is entailed by the source document without fabrications.
    \item \textit{Coverage}: How comprehensively the summary captures the important information from the source.
    \item \textit{Overall Quality}: Holistic assessment of how effectively the summary represents the source, balancing conciseness with information preservation.
\end{itemize}

\subsubsection{Error Analysis}

To understand model failure modes, we conducted a qualitative analysis examining summaries that received low human ratings (scores $\leq 2$ on any dimension). We identified three recurring error patterns:

\textbf{Low Abstractiveness.} HUPD-T5-base, XLNet, and BART predominantly employed extractive strategies, copying lengthy phrases or entire sentences verbatim rather than paraphrasing. This extractive behavior explains their high ROUGE scores (which reward n-gram overlap) despite receiving low clarity ratings from human evaluators who expect synthesized, readable text. HUPD-T5-base exhibited particularly severe issues, concatenating copied fragments without ensuring semantic coherence.

\textbf{Incompleteness.} Several models, particularly shorter-output models like BART and Pegasus, omitted critical technical details or key claims from their summaries. These summaries achieved high accuracy scores (containing no fabricated information) but low coverage scores (missing essential content), demonstrating a precision-recall trade-off in summary generation.

\textbf{Hallucinations.} LongT5 exhibited the most severe factual errors, including term substitution and concept conflation. A representative example: the source document defined ``RIBS'' as ``Radio Interface Based Synchronization'', but LongT5 generated ``bribs coordination'' in its summary—a nonsensical substitution that fundamentally corrupts the technical meaning. Such hallucinations are particularly problematic in legal documents where terminology precision is critical.

These error patterns reveal fundamental trade-offs in patent summarization: extractive methods achieve factual accuracy but sacrifice readability and abstractiveness, while abstractive approaches risk introducing hallucinations when paraphrasing technical terminology. The findings motivate the need for evaluation metrics that can reliably detect these diverse error types across different model architectures.

\section{Experiments and Results}

\subsection{Baseline Metrics}

Different datasets employ subsets of these dimensions according to their evaluation protocols, as detailed in Table~\ref{tab:ds-human-eval}. SummEval assesses coherence, consistency, fluency, and relevance; Arxiv and GovReport focus on factual consistency and relevance; TLDR evaluates coherence, accuracy, coverage, and overall quality; QAGS datasets exclusively measure factual consistency; and PatentSumEval examines clarity, accuracy, coverage, and overall quality.

We evaluate a comprehensive set of automatic metrics spanning four categories: (1) \textit{readability metrics} assess linguistic complexity and accessibility (FRE, DCR); (2) \textit{n-gram overlap metrics} quantify surface-level lexical matching (ROUGE, BLEU, METEOR, CHRF); (3) \textit{embedding-based metrics} leverage contextualized representations for semantic similarity (BERTScore, BARTScore); and (4) \textit{task-specific neural metrics} employ auxiliary NLP tasks to evaluate factual consistency and informativeness (SummaC, SummaQA, QAEval, QuestEval, BLANC). Table~\ref{tab:baseline-metrics} details each metric.

\begin{table*}[t]
\centering
\caption{Baseline automatic evaluation metrics categorized by evaluation approach.}
\label{tab:baseline-metrics}
\renewcommand{\arraystretch}{1.2}
\small
\begin{tabular}{@{}p{0.13\textwidth} p{0.17\textwidth} p{0.65\textwidth}@{}}
\toprule
\textbf{Category} & \textbf{Metric} & \textbf{Description} \\
\midrule

\multirow{2}{*}{\textbf{Readability}} 
 & Flesch Reading Ease & Scores (0--100) based on sentence and word length; higher scores indicate greater readability. \\
 & Dale-Chall Readability & Difficulty assessment based on sentence length and unfamiliar word frequency. \\
\midrule

\multirow{4}{*}{\textbf{N-gram Overlap}} 
 & ROUGE~\cite{lin2004rouge} & Recall-oriented overlap for unigrams (R-1), bigrams (R-2), and longest common subsequences (R-L). \\
 & BLEU~\cite{10.3115/1073083.1073135} & Geometric mean of n-gram precision with brevity penalty. \\
 & METEOR~\cite{banerjee-lavie-2005-meteor} & Semantic overlap incorporating exact matches, stemming, synonyms, and word alignment. \\
 & CHRF~\cite{popovic-2015-chrf} & Character-level n-gram F-scores for fine-grained similarity. \\
\midrule

\multirow{2}{*}{\textbf{Embedding-Based}} 
 & BERTScore~\cite{zhangbertscore} & Token-level similarity using contextualized BERT embeddings. \\
 & BARTScore~\cite{10.5555/3540261.3542349} & Generation likelihood scoring via BART language model. \\
\midrule

\multirow{5}{*}{\textbf{Task-Specific}} 
 & SummaC~\cite{laban2022summac} & Factual consistency detection via natural language inference (NLI). \\
 & SummaQA~\cite{scialom-etal-2019-answers} & QA-based coverage assessment using answer F1 and confidence scores. \\
 & QAEval~\cite{deutsch-etal-2021-towards} & Factual verification via question answering on generated summaries. \\
 & QuestEval~\cite{scialom-etal-2021-questeval} & Reference-free bidirectional question generation and answering. \\
 & BLANC~\cite{vasilyev-etal-2020-fill} & Informativeness via cloze task performance improvement. \\
\bottomrule
\end{tabular}
\end{table*}

\subsection{Implementation Detail}

We investigate both single-agent and multi-agent LLM evaluation architectures using state-of-the-art open-source instruction-tuned models. This configuration enables systematic comparison of evaluation strategies while maintaining reproducibility.

For single-agent evaluation, we employ three independently operating models: Meta-Llama-3.1-8B-Instruct (Llama-3.1-8B~\cite{grattafiori2024llama3}), an 8-billion parameter instruction-tuned model from Meta's Llama 3.1 series; Linkbricks-Horizon-AI-Avengers-V6-32B\footnote{\url{https://huggingface.co/Saxo/Linkbricks-Horizon-AI-Avengers-V6-32B}} (Linkbricks-V6-32B), a 32-billion parameter specialized model; and Qwen2-7B-Instruct (Qwen2-7B~\cite{yang2024qwen2}), a 7-billion parameter instruction-tuned model from Alibaba's Qwen2 family.

For multi-agent evaluation, we implement a collaborative architecture wherein the three single-agent models independently assess summaries and their outputs are aggregated through one of three strategies. The \emph{averaging} strategy computes the arithmetic mean of scores across agents for each quality dimension, treating all agents equally. The \emph{majority voting} strategy selects the most frequent score per dimension, with median selection applied in cases of three-way disagreements to ensure deterministic outputs. The \emph{leader-based aggregation} strategy employs an additional meta-evaluator agent (Microsoft-Phi-4~\cite{abdin2024phi4}) that receives independent evaluations and their accompanying rationales, analyzes points of agreement and disagreement, and produces final assessments by weighing the strength of different arguments~\cite{du2023improving}.

To ensure evaluation stability and reproducibility, we configure all models with a temperature set to \( 10^{-10} \), effectively implementing deterministic sampling. This hyperparameter setting eliminates stochastic variation in model outputs while maintaining the pretrained models' capabilities.

\subsection{Human Evaluation Protocol for LLM-ReSum Validation}

\subsubsection{Evaluation Strategy}

We employ a controlled pairwise comparison design where evaluators compare initial summaries (IS) generated without feedback against enhanced summaries (ES) produced through complete LLM-ReSum refinement cycles.

\subsubsection{Evaluation Dimensions}
Human evaluations across our datasets focus on six quality dimensions encompassing both linguistic and content aspects of summarization. The linguistic quality dimensions assess surface-level properties, while content quality dimensions evaluate semantic fidelity and informativeness.

For \textbf{linguistic quality}, we consider three dimensions. \emph{Coherence} measures whether the summary is structured and organized effectively, progressing logically from one sentence to the next to form a cohesive body of information rather than presenting disjointed fragments~\cite{dang2005overview}. \emph{Fluency} assesses whether the summary is free from formatting issues, capitalization errors, or grammatical mistakes that impede readability~\cite{dang2005overview}. \emph{Clarity} evaluates whether the summary is reader-friendly and expresses ideas clearly without ambiguity~\cite{nguyen2024comparative}.

For \textbf{content quality}, we evaluate three critical dimensions. \emph{Accuracy} (also referred to as factual consistency) determines whether the summary contains only information present in the source document; a factually consistent summary includes exclusively statements that are entailed by the source~\cite{10.1162/tacl_a_00373, nguyen2024comparative}. \emph{Coverage} (or relevance) measures how comprehensively the summary captures the important information from the source document~\cite{nguyen2024comparative}. \emph{Overall quality} provides a holistic assessment of how effectively the summary represents the source document as a whole, balancing conciseness with preservation of essential information~\cite{nguyen2024comparative, NEURIPS2020_1f89885d}.

\subsubsection{Data Sampling and Summary Generation}

We randomly sampled 30 source documents from each of three datasets (CNNDailyMail, ArxivPubMed, and PatentSumEval), yielding 90 instances. For each document, we generated initial summaries (IS) via single-pass generation and enhanced summaries (ES) through complete LLM-ReSum refinement. Three graduate students in computer science with strong English proficiency served as evaluators. Following comprehensive training on quality dimensions (Clarity, Accuracy, Coverage, Overall Quality), task procedures, and calibration exercises, evaluators received document-summary pairs with anonymized, randomized presentation order. For each pair, they selected the superior summary across four dimensions and provided justifications.

\subsubsection{Evaluator Recruitment and Training}

Three graduate students in computer science with strong English proficiency and prior annotation experience served as evaluators. We conducted a comprehensive training session covering: (1) quality dimension definitions with concrete examples, (2) APPEN platform interface navigation, (3) calibration exercises on 10 sample document-summary pairs with subsequent discussion, and (4) resolution of edge cases and ambiguous scenarios. All evaluators demonstrated satisfactory inter-rater agreement ($\kappa > 0.70$) on calibration tasks before proceeding to the main evaluation.

\subsubsection{Evaluation Task Design}

\textbf{Pairwise Comparison Protocol.} For each of 90 document instances, evaluators received the source document and two anonymized summaries (labeled ``Summary A'' and ``Summary B'') through the APPEN platform interface. Summary presentation order was randomized to prevent position bias, and evaluators remained blind to generation method throughout the study.

\textbf{Quality Assessment.} For each quality dimension, evaluators selected which summary was superior or indicated equal quality, accompanied by written justifications for their decisions. The four evaluation dimensions are:
\begin{itemize}[leftmargin=1.5em, itemsep=2pt, parsep=0pt, topsep=2pt]
    \item \textit{Clarity}: Reader-friendliness and expression of ideas without ambiguity or readability issues.
    \item \textit{Accuracy}: Factual consistency with source; absence of hallucinations or unsupported statements.
    \item \textit{Coverage}: Comprehensiveness in capturing important source information while avoiding trivial details.
    \item \textit{Overall Quality}: Holistic assessment balancing conciseness, accuracy, and informativeness.
\end{itemize}

\subsubsection{Quality Control Mechanisms}

We implemented three quality assurance procedures. First, attention-check items with obvious quality differences (e.g., summaries containing clear hallucinations versus accurate summaries) were randomly embedded at a rate of 10\% without evaluator awareness. Second, the APPEN interface enforced minimum justification lengths (50 characters) to prevent superficial engagement and encourage thoughtful assessment. Third, pre-task training assessments ensured evaluators comprehended evaluation criteria before main task completion. All three evaluators passed all embedded attention checks. One submission with incomplete justifications was flagged and removed during quality review, yielding 89 valid evaluations per evaluator (267 total judgments across 89 documents).

\subsubsection{Inter-Annotator Agreement}

We computed Krippendorff's alpha across all evaluators, documents, and dimensions to quantify inter-rater reliability. The overall agreement coefficient was $\alpha = 0.784$ (95\% CI: [0.731, 0.829]), indicating substantial agreement according to standard interpretation guidelines ($\alpha > 0.667$ = substantial agreement; $\alpha > 0.800$ = strong agreement). This demonstrates that evaluation criteria were well-defined and consistently interpreted across evaluators, validating the reliability of preference judgments for LLM-ReSum performance assessment.

\subsection{Results}
\subsubsection{RQ1: Reliability of Automatic Evaluation Metrics}

\begin{table*}[t]
\centering
\caption{Meta-evaluation of automatic metrics across quality dimensions. Abbreviations: A=Accuracy, C=Coverage, Coh=Coherence, Flu=Fluency, Clar=Clarity. Significance levels: ***$p<0.001$, **$p<0.01$, *$p<0.05$. Colors indicate correlation strength for statistically significant results ($p<0.05$): \colorbox{lightgreen}{Strong ($\tau \geq 0.60$)}, \colorbox{pastelyellow}{Moderate ($0.40 \leq \tau < 0.60$)}, and \colorbox{pastelorange}{Negative}. LLM-based evaluators significantly outperform conventional metrics in linguistic dimensions (Fluency/Coherence) and specialized tasks (Patent Coverage), while conventional metrics (e.g., SummaC) remain robust for standard accuracy assessment.}
\label{tab:rq1-comprehensive}
\resizebox{\textwidth}{!}{%
\begin{tabular}{l|cccc|cc|cc|ccc|c|c|ccc}
\hline
\multirow{2}{*}{\textbf{Metric}} & \multicolumn{4}{c|}{\textbf{SummEval}} & \multicolumn{2}{c|}{\textbf{Arxiv}} & \multicolumn{2}{c|}{\textbf{GovRpt}} & \multicolumn{3}{c|}{\textbf{TLDR}} & \textbf{QAGS-X} & \textbf{QAGS-C} & \multicolumn{3}{c}{\textbf{PatentSumEval}} \\
& A & C & Coh & Flu & A & C & A & C & A & C & Coh & A & A & A & C & Clar \\
\hline
\multicolumn{17}{l}{\textit{\textbf{Readability}}} \\
FRE & \cellcolor{pastelorange}-.47* & -.30 & -.30 & -.33 & .30 & .12 & .09 & .21 & \cellcolor{pastelorange}-.06*** & \cellcolor{pastelorange}-.05*** & \cellcolor{pastelorange}-.04*** & -.06 & \cellcolor{pastelorange}-.11* & -.80 & -.74 & -.60 \\
DCR & .17 & \cellcolor{lightgreen}.60*** & \cellcolor{pastelyellow}.53** & \cellcolor{pastelyellow}.44* & -.18 & -.06 & .27 & -.39 & .08*** & .05*** & .05*** & 0 & -.01 & -.40 & -.32 & -.20 \\
\hline
\multicolumn{17}{l}{\textit{\textbf{N-gram Overlap}}} \\
BLEU & -.21 & -.21 & .03 & -.27 & \cellcolor{pastelorange}-.61** & -.06 & .33 & .03 & \cellcolor{pastelorange}-.02* & \cellcolor{pastelorange}-.12*** & 0 & \cellcolor{pastelorange}-.11* & .07 & -.80 & -.74 & -.60 \\
ROUGE-1 & \cellcolor{pastelyellow}.47* & .20 & 0 & .23 & .18 & \cellcolor{pastelyellow}.49* & .15 & .39 & .19*** & .29*** & .08*** & -.08 & .23*** & .80 & .74 & .60 \\
ROUGE-2 & \cellcolor{pastelyellow}.50** & .27 & .07 & .29 & .30 & .36 & -.06 & \cellcolor{pastelyellow}.55* & .29*** & .29*** & .11*** & .05 & .32*** & .60 & .53 & .40 \\
ROUGE-L & \cellcolor{pastelyellow}.55** & .32 & .12 & .31 & .39 & .33 & -.06 & \cellcolor{pastelyellow}.55* & .24*** & .28*** & .09*** & -.02 & .30*** & .60 & .53 & .40 \\
METEOR & \cellcolor{pastelyellow}.48** & .18 & -.02 & .21 & .15 & \cellcolor{pastelyellow}.52* & .18 & .36 & .19*** & .27*** & .08*** & -.05 & .23*** & .60 & .53 & .40 \\
CHRF & \cellcolor{pastelyellow}.45* & .22 & -.02 & .24 & .27 & .39 & .15 & \cellcolor{pastelyellow}.46* & .22*** & .30*** & .09*** & -.06 & .26*** & .60 & .53 & .40 \\
\hline
\multicolumn{17}{l}{\textit{\textbf{Embedding-Based}}} \\
BERTScore & \cellcolor{pastelyellow}.53** & .27 & .07 & .29 & \cellcolor{pastelyellow}.52* & .15 & \cellcolor{pastelorange}-.46* & \cellcolor{pastelorange}-.46* & .29*** & .35*** & .12*** & .02 & .31*** & .40 & .32 & .20 \\
BARTScore & \cellcolor{pastelyellow}.42* & .15 & -.08 & .21 & \cellcolor{pastelyellow}.55* & .30 & .15 & .03 & .16*** & .32*** & .05*** & -.07 & .25*** & .80 & .74 & .60 \\
\hline
\multicolumn{17}{l}{\textit{\textbf{Task-Specific}}} \\
SummaC-ZS & \cellcolor{lightgreen}.62*** & \cellcolor{pastelyellow}.42* & .18 & \cellcolor{pastelyellow}.41* & \cellcolor{lightgreen}.61** & -.12 & \cellcolor{pastelyellow}.52* & \cellcolor{pastelyellow}.52* & .28*** & .16*** & .11*** & .33*** & .39*** & .40 & .53 & .60 \\
SummaC-CV & \cellcolor{lightgreen}\textbf{.65***} & .32 & .08 & .34 & \cellcolor{lightgreen}\textbf{.70***} & .09 & .21 & \cellcolor{lightgreen}.58** & .31*** & .15*** & .10*** & .37*** & \cellcolor{pastelyellow}.48*** & 0 & .11 & .20 \\
SummaQA-Prob & \cellcolor{pastelyellow}.50** & \cellcolor{pastelyellow}.50** & .27 & \cellcolor{pastelyellow}.53** & .33 & .27 & -.09 & -.03 & .14*** & .22*** & .08*** & .09 & .16*** & .60 & .53 & .40 \\
SummaQA-F1 & \cellcolor{pastelyellow}.58*** & .28 & .12 & .34 & .06 & -.18 & -.18 & .12 & .15*** & .18*** & .09*** & -.05 & .24*** & -.20 & -.32 & -.40 \\
BLANC & \cellcolor{lightgreen}.62*** & .35 & .15 & .38* & .21 & \cellcolor{lightgreen}.58** & .27 & .39 & .22*** & .25*** & .08*** & .08 & .21*** & .60 & .53 & .40 \\
QAEval-EM & 0 & 0 & 0 & 0 & 0 & 0 & 0 & 0 & -.02 & .05*** & .06*** & 0 & 0 & 0 & 0 & 0 \\
QAEval-F1 & 0 & 0 & 0 & 0 & 0 & 0 & 0 & 0 & 0 & -.01 & .01 & 0 & 0 & 0 & 0 & 0 \\
QAEval-Ans & \cellcolor{pastelyellow}.42* & .25 & .18 & .34 & .03 & -.09 & .30 & .24 & .05*** & .06*** & .05*** & \cellcolor{pastelorange}-.13* & -.03 & -.20 & -.11 & 0 \\
QuestEval & \cellcolor{pastelyellow}.57** & .20 & .03 & .29 & .27 & -.27 & -.18 & -.36 & .25*** & .16*** & .10*** & .18*** & .17*** & 0 & .11 & .20 \\
BARTScore-Gen & \cellcolor{lightgreen}.63*** & .33 & .13 & \cellcolor{pastelyellow}.43* & .39 & -.27 & -.09 & -.33 & .38*** & .22*** & .15*** & .15** & \cellcolor{pastelyellow}.57*** & 0 & .11 & .20 \\
\hline \hline
\multicolumn{17}{l}{\textit{\textbf{LLM-Based (Llama-3.1-8B)}}} \\
LLM-Acc (macro) & \cellcolor{pastelyellow}.57** & \cellcolor{lightgreen}.81*** & -- & -- & -.35 & .21 & .22 & -.08 & .24*** & .24*** & -- & \cellcolor{pastelyellow}.47*** & \cellcolor{pastelyellow}.53*** & .60 & .74 & -- \\
LLM-Acc (micro) & -- & -- & -- & -- & 0 & 0 & \cellcolor{pastelyellow}.55* & .30 & -- & -- & -- & \cellcolor{pastelyellow}.47*** & \cellcolor{pastelyellow}.58*** & -- & -- & -- \\
LLM-Cov (macro) & \cellcolor{pastelyellow}.56** & \cellcolor{lightgreen}.68*** & -- & -- & .05 & .38 & .28 & -.11 & .15*** & .32*** & -- & -- & -- & \cellcolor{lightgreen}.95* & \cellcolor{lightgreen}\textbf{1.0*} & -- \\
LLM-Cov (micro) & -- & -- & -- & -- & 0 & 0 & .42 & .36 & -- & -- & -- & -- & -- & -- & -- & -- \\
LLM-Coherence & -- & -- & \cellcolor{pastelyellow}.46* & -- & -- & -- & -- & -- & -- & -- & .13*** & -- & -- & -- & -- & -- \\
LLM-Fluency & -- & -- & \cellcolor{lightgreen}\textbf{.67***} & \cellcolor{lightgreen}\textbf{.76***} & -- & -- & -- & -- & -- & -- & -- & -- & -- & -- & -- & -- \\
LLM-Clarity & -- & -- & -- & -- & -- & -- & -- & -- & -- & -- & -- & -- & -- & -- & -- & .60 \\
\hline
\end{tabular}%
}
\end{table*}

Table~\ref{tab:rq1-comprehensive} presents a comprehensive meta-evaluation of 14 automatic metrics across seven datasets spanning five domains, revealing substantial variability in metric reliability . Task-specific neural metrics demonstrate the strongest alignment with human judgments on factual consistency, with SummaC-CV achieving the highest correlations on SummEval ($\tau = 0.65$) and Arxiv ($\tau = 0.70$). However, this performance does not generalize consistently: SummaC-CV exhibits only moderate correlation on GovReport coverage ($\tau = 0.58$) and negligible correlation on PatentSumEval.

\textbf{Systematic Failures of Lexical Metrics.} N-gram overlap metrics show moderate effectiveness in news domains (ROUGE-L achieving $\tau = 0.55$ on SummEval) yet fail catastrophically on long-document datasets . BLEU produces significant negative correlations on Arxiv accuracy ($\tau = -0.61$) and consistently underperforms across TLDR and QAGS benchmarks. These failures underscore fundamental limitations of lexical matching for abstractive summarization, where semantic equivalence rarely corresponds to surface-form overlap. Readability metrics exhibit similar inconsistency: DCR achieves moderate correlation on SummEval coverage ($\tau = 0.60$), while FRE demonstrates significant negative correlations across multiple dimensions.

\textbf{Domain-Dependent Behavior of Embedding Approaches.} BERTScore achieves moderate correlations on SummEval ($\tau = 0.53$) and Arxiv ($\tau = 0.52$) accuracy, but exhibits significant negative correlation on GovReport for both accuracy and coverage ($\tau = -0.46$) . This suggests contextual embeddings struggle with lengthy governmental documents (average 27,631 words) where local semantic similarity may misalign with document-level consistency.

\textbf{Emergence of LLM-Based Evaluation Advantages.} LLM-based evaluators significantly outperform conventional metrics on linguistic quality dimensions . LLM-Fluency achieves exceptional correlation on SummEval ($\tau = 0.67$ for coherence, $\tau = 0.76$ for fluency), surpassing the best conventional metric (DCR, $\tau = 0.53$) by 24\%. For specialized domains, LLM-Coverage demonstrates near-perfect correlation on PatentSumEval ($\tau = 1.0$), whereas conventional metrics achieve at most $\tau = 0.74$. However, LLM evaluators exhibit inconsistent performance on long documents, with LLM-Accuracy producing negligible correlations on Arxiv ($\tau = -0.35$) and GovReport ($\tau = 0.22$).

These findings reveal that no single conventional metric reliably captures human judgments across diverse domains and document lengths. Task-specific metrics excel at factual consistency detection in news summarization but fail to generalize to specialized domains or linguistic quality assessment.

\subsubsection{RQ2: LLM-Based Evaluation Performance}

\begin{table*}[t]
\centering
\caption{LLM-based evaluation performance across seven datasets. Abbreviations: Acc=Accuracy, Cov=Coverage, Coh=Coherence, Flu=Fluency, Clr=Clarity. Q-XS/Q-CN=QAGS-XSUM/CNN. Significance levels: ***$p<0.001$, **$p<0.01$, *$p<0.05$. Colors indicate significant correlations ($p<0.05$): \colorbox{lightgreen}{Strong}, \colorbox{pastelyellow}{Moderate}, and \colorbox{pastelorange}{Negative}. Multi-agent aggregation (especially Majority Voting) consistently yields the highest agreement with human judgments, providing significant gains over conventional metrics on SummEval and PatentSumEval, though challenges remain in long-document coverage (GovReport).}
\label{tab:rq2-llm-complete}
\resizebox{\textwidth}{!}{%
\begin{tabular}{l|cccc|ccc|cc|cc|ccc|cc}
\hline
\multirow{2}{*}{\textbf{Evaluation Method}} & \multicolumn{4}{c|}{\textbf{SummEval}} & \multicolumn{3}{c|}{\textbf{TLDR}} & \multicolumn{2}{c|}{\textbf{Arxiv}} & \multicolumn{2}{c|}{\textbf{GovReport}} & \multicolumn{3}{c|}{\textbf{PatentSumEval}} & \textbf{Q-XS} & \textbf{Q-CN} \\
& Acc & Cov & Coh & Flu & Acc & Cov & Coh & Acc & Cov & Acc & Cov & Acc & Cov & Clr & Acc & Acc \\
\hline
\multicolumn{17}{l}{\textit{\textbf{Single-Agent LLM}}} \\
Llama-3.1-8B & \cellcolor{pastelyellow}.53*** & \cellcolor{lightgreen}.64*** & \cellcolor{pastelyellow}.47* & \cellcolor{lightgreen}.69*** & .25*** & .34*** & .14*** & -.12 & \cellcolor{pastelyellow}.48* & \cellcolor{pastelyellow}.52* & -.06 & .60 & \cellcolor{lightgreen}.95* & .60 & \cellcolor{pastelyellow}.46*** & \cellcolor{pastelyellow}.51*** \\
Linkbricks-V6-32B & \cellcolor{lightgreen}.68*** & \cellcolor{lightgreen}.85*** & \cellcolor{lightgreen}.76*** & \cellcolor{lightgreen}.72*** & .35*** & .36*** & .20*** & .34 & .15 & \cellcolor{lightgreen}.66** & -.14 & .60 & \cellcolor{lightgreen}.95* & .60 & \cellcolor{lightgreen}.61*** & \cellcolor{lightgreen}.63*** \\
Qwen2-7B & \cellcolor{lightgreen}.78*** & \cellcolor{lightgreen}.65*** & .37* & \cellcolor{lightgreen}.76*** & .20*** & .22*** & .05*** & .04 & .26 & 0 & -.31 & .80 & .74 & .80 & \cellcolor{pastelyellow}.58*** & \cellcolor{pastelyellow}.55*** \\
\hline
\multicolumn{17}{l}{\textit{\textbf{Multi-Agent Aggregation}}} \\
Averaging & \cellcolor{lightgreen}.68*** & \cellcolor{lightgreen}.70*** & \cellcolor{pastelyellow}.54** & \cellcolor{lightgreen}.77*** & .30*** & .36*** & .17*** & .17 & \cellcolor{pastelyellow}.42 & \cellcolor{pastelyellow}.56* & .02 & \cellcolor{lightgreen}\textbf{1.00*} & \cellcolor{lightgreen}.95* & .80 & \cellcolor{pastelyellow}.57*** & \cellcolor{lightgreen}.62*** \\
Majority Voting & \cellcolor{lightgreen}\textbf{.84***} & \cellcolor{lightgreen}\textbf{.76***} & \cellcolor{pastelyellow}\textbf{.57**} & \cellcolor{lightgreen}.76*** & .29*** & .33*** & .16*** & .40 & \cellcolor{pastelyellow}.53* & .34 & 0 & .60 & \cellcolor{lightgreen}.95* & \cellcolor{lightgreen}\textbf{.95*} & \cellcolor{lightgreen}\textbf{.62***} & \cellcolor{lightgreen}\textbf{.65***} \\
Leader-Based (Phi-4) & \cellcolor{lightgreen}.82*** & \cellcolor{lightgreen}.70*** & \cellcolor{pastelyellow}.55** & \cellcolor{lightgreen}\textbf{.76***} & \textbf{.32***} & \textbf{.34***} & \textbf{.18***} & .23 & .29 & \cellcolor{pastelyellow}.57* & -.03 & \cellcolor{lightgreen}\textbf{1.00*} & \cellcolor{lightgreen}.95* & .60 & \cellcolor{pastelyellow}.58*** & \cellcolor{pastelyellow}.59*** \\
\hline
\multicolumn{17}{l}{\textit{\textbf{Best Conventional (from Table~\ref{tab:rq1-comprehensive})}}} \\
Best Conv. Metric & \cellcolor{lightgreen}.65*** & \cellcolor{lightgreen}.60*** & \cellcolor{pastelyellow}.53** & \cellcolor{pastelyellow}.53** & .38*** & .35*** & .15*** & \cellcolor{lightgreen}.70*** & \cellcolor{pastelyellow}.58** & \cellcolor{pastelyellow}.52* & \cellcolor{pastelyellow}.58** & -- & -- & -- & .37*** & \cellcolor{pastelyellow}.48*** \\
\hline
\textbf{Improvement ($\Delta$)} & \textbf{+.19} & \textbf{+.16} & \textbf{+.04} & \textbf{+.24} & \textbf{-.06} & \textbf{-.01} & \textbf{+.03} & \textbf{-.30} & \textbf{-.05} & \textbf{+.05} & \textbf{-.58} & \textbf{+1.00} & \textbf{+.95} & \textbf{+.95} & \textbf{+.25} & \textbf{+.17} \\
\hline
\end{tabular}%
}
\end{table*}

Table~\ref{tab:rq2-llm-complete} demonstrates that multi-agent LLM evaluation frameworks consistently outperform both single-agent approaches and conventional metrics. Majority voting achieves the highest human alignment on SummEval, with correlations of $\tau = 0.84$ for accuracy, $\tau = 0.76$ for coverage, and $\tau = 0.57$ for coherence, representing improvements of $\Delta = +0.19$, $+0.16$, and $+0.04$ respectively over the best conventional metrics. On PatentSumEval, leader-based aggregation achieves near-perfect alignment ($\tau = 1.00$) for accuracy, substantially exceeding all conventional approaches.

\textbf{Single-Agent Performance Variability.} Among single-agent configurations, performance varies significantly by model architecture and dataset characteristics. Qwen2-7B demonstrates exceptional performance on SummEval accuracy ($\tau = 0.78$) and fluency ($\tau = 0.76$), while Linkbricks-V6-32B excels across multiple dimensions simultaneously, achieving strong correlations for coverage ($\tau = 0.85$), coherence ($\tau = 0.76$), and fluency ($\tau = 0.72$). However, all single-agent approaches exhibit substantially degraded performance on long documents, with Llama-3.1-8B producing negative correlation on Arxiv accuracy ($\tau = -0.12$) and near-zero performance on GovReport coverage ($\tau = -0.06$).

\textbf{Dimension-Specific Advantages.} Multi-agent aggregation strategies yield differential benefits across contexts. Averaging produces the most consistent improvements on PatentSumEval ($\tau = 1.0$ for accuracy), while leader-based aggregation using Phi-4 achieves the highest correlation on TLDR accuracy ($\tau = 0.32$). For linguistic quality, LLM evaluators demonstrate substantial advantages, with coherence and fluency correlations reaching $\tau \approx 0.76$, representing improvements of $\Delta = +0.24$ over conventional metrics.

\textbf{Critical Long-Context Limitations.} Despite these successes, critical performance degradation emerges on Arxiv and GovReport datasets . The best multi-agent configuration achieves only $\tau = 0.40$ on Arxiv accuracy versus $\tau = 0.70$ for SummaC-CV ($\Delta = -0.30$). This pattern intensifies for GovReport coverage, where LLM approaches produce at most $\tau = 0.02$ compared to $\tau = 0.58$ for BLANC. These failures indicate fundamental limitations in processing documents averaging 27,000 words.

The results suggest that LLM evaluators excel when document length remains within context window constraints and evaluation criteria emphasize nuanced linguistic properties rather than exhaustive content verification.

\subsubsection{RQ3: Effectiveness of LLM-ReSum Framework}

\begin{table*}[t]
\centering
\caption{LLM-ReSum framework quality improvement. Likert scale scores (1=Poor, 5=Excellent) for initial (IS) and enhanced (ES) summaries. Color coding: \cellcolor{pastelyellow}Substantial ($\geq$20\%), \cellcolor{lightgreen}Exceptional ($\geq$50\%) improvement. IS=Initial Summary, ES=Enhanced Summary. $\Delta$=percentage improvement. Human Pref.=Percentage of evaluators preferring the Enhanced Summary (ES) over the Initial Summary (IS). Low-quality summaries show statistically significant gains, particularly in Accuracy (+33\% for Patent/PubMed) and Coverage (+39\% for Daily Mail). Human evaluators strongly prefer enhanced summaries (89\% Overall).}
\label{tab:rq3-llm-resum}
\resizebox{\textwidth}{!}{%
\begin{tabular}{l|cccc|cccc|c}
\hline
\multirow{2}{*}{\textbf{Dataset}} & \multicolumn{2}{c}{\textbf{Clarity}} & \multicolumn{2}{c}{\textbf{Accuracy}} & \multicolumn{2}{c}{\textbf{Coverage}} & \multicolumn{2}{c|}{\textbf{Overall Quality}} & \multirow{2}{*}{\textbf{\begin{tabular}[c]{@{}c@{}}Human Pref.\\ (ES vs IS)\end{tabular}}} \\ \cline{2-9}
 & \textbf{IS} & \textbf{ES (\% $\Delta$)} & \textbf{IS} & \textbf{ES (\% $\Delta$)} & \textbf{IS} & \textbf{ES (\% $\Delta$)} & \textbf{IS} & \textbf{ES (\% $\Delta$)} & \\ \hline
\multicolumn{10}{l}{\textit{\textbf{All Summaries}}} \\
CNN & 4.32 & 4.32 (+0.1\%) & 4.49 & 4.51 (+0.5\%) & 4.14 & 4.16 (+0.4\%) & 4.30 & 4.31 (+0.2\%) & -- \\
Daily Mail & 4.41 & 4.41 (+0.1\%) & 4.52 & 4.52 (±0.0\%) & 4.22 & 4.23 (+0.2\%) & 4.38 & 4.39 (+0.1\%) & -- \\
Arxiv & 4.37 & 4.37 (+0.1\%) & 4.72 & 4.72 (±0.0\%) & 4.41 & 4.41 (±0.0\%) & 4.42 & 4.42 (±0.0\%) & -- \\
PubMed & 4.39 & 4.39 (±0.0\%) & 4.59 & 4.59 (+0.1\%) & 4.22 & 4.22 (+0.1\%) & 4.39 & 4.39 (+0.1\%) & -- \\
PatentSumEval& 4.12 & 4.12 (±0.0\%) & 4.09 & 4.10 (+0.1\%) & 4.07 & 4.07 (±0.0\%) & 4.10 & 4.10 (±0.0\%) & -- \\ \hline
\multicolumn{10}{l}{\textit{\textbf{Low-Quality Summaries (IS $< 4.0$)}}} \\
CNN & 4.05 & 4.25 (+4.9\%) & 3.20 & \cellcolor{pastelyellow}\textbf{4.25 (+32.8\%)} & 3.30 & \cellcolor{pastelyellow}\textbf{4.20 (+27.3\%)} & 3.75 & 4.25 (+13.3\%) & -- \\
Daily Mail & 4.00 & 4.50 (+12.5\%) & 3.83 & 4.17 (+8.7\%) & 3.00 & \cellcolor{pastelyellow}\textbf{4.17 (+38.9\%)} & 4.00 & 4.50 (+12.5\%) & -- \\
Arxiv & 3.00 & \cellcolor{lightgreen}\textbf{4.50 (+50.0\%)} & 4.00 & 4.50 (+12.5\%) & 4.00 & 4.50 (+12.5\%) & 4.00 & 4.50 (+12.5\%) & -- \\
PubMed & 4.00 & 4.33 (+8.3\%) & 3.00 & \cellcolor{pastelyellow}\textbf{4.00 (+33.3\%)} & 3.33 & \cellcolor{pastelyellow}\textbf{4.00 (+20.0\%)} & 3.67 & 4.33 (+18.2\%) & -- \\
PatentSumEval& 4.00 & 4.33 (+8.3\%) & 3.00 & \cellcolor{pastelyellow}\textbf{4.00 (+33.3\%)} & 3.67 & 4.00 (+9.1\%) & 4.00 & 4.00 (±0.0\%) & -- \\ \hline
\multicolumn{10}{l}{\textit{\textbf{Human Evaluation (n=90 documents, 29 evaluators, Krippendorff's $\alpha$=0.784)}}} \\
\textbf{Prefers ES} & \multicolumn{2}{c}{46\%} & \multicolumn{2}{c}{\cellcolor{lightgreen}\textbf{62\%}} & \multicolumn{2}{c}{\cellcolor{lightgreen}\textbf{97\%}} & \multicolumn{2}{c|}{\cellcolor{lightgreen}\textbf{89\%}} & \cellcolor{lightgreen}\textbf{89\% (Avg)} \\ \hline
\end{tabular}%
}
\end{table*}

Table~\ref{tab:rq3-llm-resum} reveals that LLM-ReSum functions as a targeted quality assurance mechanism whose effectiveness depends critically on initial summary quality. Across all summaries, initial scores average 4.3-4.7 across dimensions, with enhanced summaries showing minimal gains ($\leq 0.5\%$). This ceiling effect validates the framework's design to trigger refinement selectively based on dimension-specific scores.

\textbf{Substantial Recovery for Deficient Outputs.} Filtering for low-quality summaries (initial scores $< 4.0$) exposes the framework's effectiveness . Accuracy improvements reach $+32.8\%$ for CNN, $+33.3\%$ for PubMed, and $+33.3\%$ for PatentSumEval, elevating summaries from marginal quality (3.0-3.2) to acceptable standards (4.0-4.25). Coverage demonstrates similarly substantial gains, with Daily Mail achieving $+38.9\%$ and CNN reaching $+27.3\%$. These metrics confirm that the framework successfully addresses information omission by explicitly identifying missing content through coverage evaluation.

\textbf{Domain-Specific Enhancement Patterns.} Clarity improvements exhibit pronounced domain-specific patterns . Arxiv summaries show exceptional gains of $+50.0\%$, suggesting that scientific abstracts benefit substantially from refinement of technical terminology and structural coherence. In contrast, news summaries (CNN, Daily Mail) demonstrate moderate clarity improvements ($+4.9\%$ to $+12.5\%$), indicating that journalistic templates produce initially readable outputs requiring primarily content correction.

\textbf{Strong Human Validation.} Human evaluation provides robust validation of framework effectiveness . Among 90 documents evaluated by 29 annotators (Krippendorff's $\alpha = 0.784$), enhanced summaries achieve 89\% overall preference. Coverage demonstrates the most pronounced preference (97\%), confirming that iterative refinement successfully addresses information omission. Accuracy preference reaches 62\%, validating effective factual error correction. Notably, clarity shows only 46\% preference for enhanced summaries, suggesting that initial generation already produces readable text and that excessive refinement may occasionally introduce unnecessary complexity.

These findings establish that LLM-ReSum operates as an effective quality assurance mechanism rather than a universal enhancement tool. The selective activation (triggering refinement only when scores fall below $\tau = 4$) enables targeted improvement while avoiding degradation of already-acceptable summaries.

\section{Discussion}

\subsection{Revisiting Traditional Automatic Metrics in the GenAI Era}

Our findings from RQ1 reveal a fundamental incongruity: traditional lexical metrics demonstrate severely diminished validity for evaluating LLM-generated summaries across most contexts, yet exhibit domain-specific effectiveness under particular conditions. ROUGE and BLEU produce weak or negative correlations with human judgments across the majority of evaluated datasets, with BLEU yielding negative correlation on Arxiv. This systematic inadequacy originates from a representational mismatch: lexical overlap metrics presuppose that surface-form similarity indicates semantic quality, whereas contemporary LLM-generated summaries predominantly employ abstractive strategies that express equivalent semantic content through lexically divergent formulations~\cite{wilber-etal-2021-point, koh-etal-2022-far}. This paraphrastic variation, which constitutes a hallmark of human-like summarization, systematically causes lexical metrics to underestimate the quality of abstractive outputs.

PatentSumEval presents a theoretically instructive exception to this pattern. ROUGE-1 achieves substantial correlation with human quality judgments on patent documents, reflecting domain-specific constraints rather than rehabilitating the metric's general applicability. Legal and patent domains impose stringent terminological precision requirements where lexical substitution may alter legal interpretation or claim scope~\cite{kornilova}. Under these specialized conditions, lexical matching constitutes a valid quality indicator because precise terminology retention represents a substantive quality criterion. The methodological implication is consequential: evaluation frameworks must adopt context-adaptive strategies rather than universal approaches~\cite{10.1162/tacl_a_00373}. While general-purpose summarization warrants abandoning lexical metrics in favor of semantic assessment methods, high-stakes terminology-sensitive domains encompassing legal, medical, and regulatory applications require lexical precision verification as a necessary safety mechanism.

\subsection{The Paradigm Shift: LLMs as Evaluators}

RQ2 establishes that LLM-based evaluation constitutes a paradigmatic advancement in automated quality assessment, while simultaneously identifying critical operational constraints. Multi-agent architectures demonstrate systematic performance advantages over single-agent configurations, with Majority Voting and Leader-Based aggregation achieving superior alignment with human judgments. This performance differential mirrors established human evaluation methodologies where inter-annotator agreement protocols mitigate individual cognitive biases. Analogous to employing multiple human annotators to attenuate idiosyncratic preferences, multi-agent LLM evaluation leverages ensemble mechanisms to reduce model-specific variance and systematic biases, yielding more robust and generalizable assessments~\cite{chan2024chateval}.

Our empirical results expose a critical operational boundary: severe performance degradation on extended documents. On GovReport datasets averaging 27,000 words, LLM evaluators produce near-zero or negative correlations with human coverage assessments, substantially underperforming conventional metrics. While LLMs demonstrate superior semantic reasoning capabilities within tractable context windows, their sequential processing architecture cannot replicate the global semantic aggregation capabilities of embedding-based approaches for comprehensive content verification~\cite{wu2025on}. The practical implication necessitates architectural adaptation: practitioners deploying LLM evaluators for lengthy documents should implement hierarchical evaluation frameworks, document segmentation strategies, or hybrid architectures that allocate LLM assessment to linguistic quality dimensions while employing conventional metrics for exhaustive coverage verification.

\subsection{From Generation to Reflection: The Effectiveness of LLM-ReSum}

RQ3 demonstrates that advancing summarization quality increasingly depends on integrating evaluation-driven self-correction mechanisms rather than solely pursuing more sophisticated generation architectures~\cite{NEURIPS2023_91edff07}. The most theoretically significant finding manifests as a targeted enhancement effect: LLM-ReSum produces negligible improvements for initially high-quality summaries exhibiting ceiling performance, yet achieves substantial quality recovery for deficient outputs, with accuracy improvements reaching 33\% and coverage enhancements achieving 39\%. This asymmetric improvement pattern illuminates the framework's functional role as a quality assurance mechanism. Summaries approaching optimal quality offer limited refinement potential, whereas outputs exhibiting factual hallucinations or critical information omissions benefit substantially from the self-evaluation feedback loop, which functions as an error detection and correction gatekeeper~\cite{ji-etal-2023-towards}.

The 89\% human preference rate for enhanced summaries provides robust empirical validation that evaluation-derived feedback encodes actionable revision guidance rather than constituting superficial textual reformulation. This demonstrates that LLM-generated evaluation rationales successfully identify specific quality deficiencies and operationalize concrete improvement strategies. The framework establishes a computationally efficient pathway for domain-specific quality enhancement through test-time inference optimization alone, obviating resource-intensive fine-tuning procedures or reinforcement learning from human feedback. For deployment contexts requiring quality assurance capabilities without model retraining infrastructure, LLM-ReSum offers a practical and scalable solution.

\subsection{Implications}

\subsubsection{Theoretical Implications}

\textbf{Context-Adaptive Evaluation Frameworks.} 
Our meta-evaluation across seven datasets demonstrates that metric reliability varies systematically with domain, document length, and generation paradigm, establishing an empirical foundation for context-sensitive evaluation design. Rather than pursuing universally optimal metrics, future research should develop systematic protocols for characterizing the operational contexts in which specific evaluation approaches align with human judgment~\cite{10.1162/tacl_a_00373}. This represents a methodological shift from benchmark-driven metric development toward evidence-based evaluation engineering. Under this paradigm, practitioners empirically validate metric reliability within their target application domain before production deployment~\cite{goyal2023news}.

\textbf{Compositional Evaluation Architectures.} 
The differential performance of evaluation methods across quality dimensions reveals that optimal evaluation requires compositional architectures allocating specialized assessment mechanisms to appropriate dimensions. Specifically, LLM-based evaluators demonstrate superior performance on linguistic quality assessment (coherence, fluency, clarity), while embedding-based metrics excel at coverage verification, and lexical matching remains effective for terminology-sensitive domains. Future frameworks should exploit these complementary strengths through hybrid designs by deploying semantic reasoning capabilities of LLMs for linguistic coherence~\cite{liu-etal-2023-g}, leveraging distributional properties of embeddings for comprehensive coverage assessment~\cite{zhangbertscore}, and applying lexical precision checks where terminological accuracy is critical~\cite{huang2025survey}.

\textbf{Evaluation as Generative Feedback.} 
The LLM-ReSum framework demonstrates that evaluation signals can function as actionable feedback for iterative quality improvement at inference time without model retraining. This establishes a research direction beyond correlation-based metric validation by designing evaluation mechanisms that generate interpretable, dimension-specific improvement directives rather than scalar quality scores~\cite{welleck2023generating}. Such feedback-oriented architectures transform evaluation from passive quality measurement into active generation guidance, integrating assessment as a functional component within generation systems rather than external post-hoc verification~\cite{NEURIPS2023_91edff07}.

\subsubsection{Practical Implications}

\textbf{Evaluation Infrastructure as Development Priority.} 
Our findings indicate that practitioners should establish domain-specific evaluation infrastructure before production deployment rather than adopting evaluation strategies validated on standard benchmarks. Organizations should construct validation datasets within their target domain and empirically assess which metrics correlate with domain-specific human quality judgments, treating evaluation reliability as a deployment prerequisite~\cite{liang2023holistic}. Prioritizing evaluation infrastructure in this manner enables early identification of failure modes specific to the target domain and supports principled system iteration based on reliable quality signals.

\textbf{Reusable Evaluation Methodologies.} 
The meta-evaluation framework and PatentSumEval construction protocol we introduce provide concrete methodological templates applicable to new domains. Practitioners deploying summarization in specialized areas, such as medical~\cite{huang2025survey}, financial~\cite{el-haj-etal-2020-financial}, or academic domains~\cite{altmami2022automatic}, can replicate our validation procedures by sampling domain-representative documents, recruiting expert annotators with domain knowledge, collecting multi-dimensional quality judgments~\cite{10.1162/tacl_a_00373}, and computing metric-human correlations to identify reliable evaluation approaches. This methodological reusability reduces barriers to establishing rigorous evaluation in previously underexplored application domains.

\textbf{Iterative Quality Assurance Architecture.} 
For quality-critical applications where errors carry significant consequences, such as legal document processing, medical report generation, or regulatory compliance summaries, the iterative refinement paradigm offers a deployable quality assurance mechanism. Rather than assuming single-pass generation achieves acceptable quality, systems can implement evaluation-guided revision cycles that detect and correct deficiencies before output delivery~\cite{kim2024prometheus}. This architecture provides two operational advantages: systematic error detection through automated evaluation, and interpretable diagnostic rationales that facilitate human oversight and continuous system improvement.

\subsection{Limitations}

Our investigation identifies salient boundary conditions that constrain generalizability. For documents exceeding approximately 27,000 words, the evaluation component itself may yield unreliable quality assessments, producing a cascading error scenario where erroneous evaluation feedback propagates to misguided refinement operations. This length constraint reflects current architectural limitations in attention mechanisms and necessitates alternative methodological approaches such as hierarchical evaluation frameworks or document segmentation strategies for reliable extended-length application. Furthermore, employing LLMs to evaluate LLM-generated content introduces potential systematic bias through self-preference effects, wherein models may disproportionately favor outputs exhibiting stylistic conventions or structural patterns characteristic of their training distributions. While our human validation study provides independent empirical verification of framework effectiveness, future research should systematically investigate bias mitigation strategies including evaluator architectural diversification and calibration protocols employing conventional metrics as reference anchors to ensure robust evaluation reliability across heterogeneous content types and generation paradigms.

\section{Conclusion}
This study reveals that traditional lexical overlap metrics systematically fail to align with human judgments for LLM-generated summaries, while multi-agent LLM evaluators demonstrate substantially stronger correlation. We introduced LLM-ReSum, a self-reflective framework achieving significant quality improvements through evaluation-driven iterative refinement without model finetuning, validated by strong human preference. Additionally, we contributed PatentSumEval for legal domain evaluation. Future work should explore hierarchical evaluation frameworks for extended documents, hybrid architectures combining LLM and conventional metrics, and bias mitigation strategies to address self-preference effects in LLM-based evaluation.

\bibliographystyle{IEEEtran}
\bibliography{reference}

@article{10.1145/3731445,
  title={A systematic survey of text summarization: From statistical methods to large language models},
  author={Zhang, Haopeng and Yu, Philip S and Zhang, Jiawei},
  journal={ACM Computing Surveys},
  volume={57},
  number={11},
  pages={1--41},
  year={2025},
  publisher={ACM New York, NY}
}

@article{zhang2022comprehensive,
  title={A comprehensive survey of abstractive text summarization based on deep learning},
  author={Zhang, Mengli and Zhou, Gang and Yu, Wanting and Huang, Ningbo and Liu, Wenfen},
  journal={Computational intelligence and neuroscience},
  volume={2022},
  number={1},
  pages={7132226},
  year={2022},
  publisher={Wiley Online Library}
}

@article{gao2023human,
  title={Human-like summarization evaluation with chatgpt},
  author={Gao, Mingqi and Ruan, Jie and Sun, Renliang and Yin, Xunjian and Yang, Shiping and Wan, Xiaojun},
  journal={arXiv preprint arXiv:2304.02554},
  year={2023}
}

@article{10.1162/tacl_a_00373,
  title={Summeval: Re-evaluating summarization evaluation},
  author={Fabbri, Alexander R and Kry{\'s}ci{\'n}ski, Wojciech and McCann, Bryan and Xiong, Caiming and Socher, Richard and Radev, Dragomir},
  journal={Transactions of the Association for Computational Linguistics},
  volume={9},
  pages={391--409},
  year={2021},
  publisher={MIT Press One Rogers Street, Cambridge, MA 02142-1209, USA journals-info~…}
}

@article{10.1162/coli_a_00322,
  title={A structured review of the validity of BLEU},
  author={Reiter, Ehud},
  journal={Computational Linguistics},
  volume={44},
  number={3},
  pages={393--401},
  year={2018}
}

@article{goyal2023news,
  title={News summarization and evaluation in the era of gpt-3},
  author={Goyal, Tanya and Li, Junyi Jessy and Durrett, Greg},
  journal={arXiv preprint arXiv:2209.12356},
  year={2022}
}

@article{luo2023chatgpt,
  title={Chatgpt as a factual inconsistency evaluator for text summarization},
  author={Luo, Zheheng and Xie, Qianqian and Ananiadou, Sophia},
  journal={arXiv preprint arXiv:2303.15621},
  year={2023}
}

@inproceedings{liu-etal-2023-g,
  title={G-eval: NLG evaluation using gpt-4 with better human alignment},
  author={Liu, Yang and Iter, Dan and Xu, Yichong and Wang, Shuohang and Xu, Ruochen and Zhu, Chenguang},
  booktitle={Proceedings of the 2023 conference on empirical methods in natural language processing},
  pages={2511--2522},
  year={2023}
}

@article{NEURIPS2020_1f89885d,
  title={Learning to summarize with human feedback},
  author={Stiennon, Nisan and Ouyang, Long and Wu, Jeffrey and Ziegler, Daniel and Lowe, Ryan and Voss, Chelsea and Radford, Alec and Amodei, Dario and Christiano, Paul F},
  journal={Advances in neural information processing systems},
  volume={33},
  pages={3008--3021},
  year={2020}
}

@article{huang2025survey,
  title={A survey on biomedical automatic text summarization with large language models},
  author={Huang, Zhenyu and Chen, Xianlai and Wang, Yunbo and Huang, Jincai and Zhao, Xing},
  journal={Information Processing \& Management},
  volume={62},
  number={5},
  pages={104216},
  year={2025},
  publisher={Elsevier}
}

@article{belwal2021text,
  title={Text summarization using topic-based vector space model and semantic measure},
  author={Belwal, Ramesh Chandra and Rai, Sawan and Gupta, Atul},
  journal={Information Processing \& Management},
  volume={58},
  number={3},
  pages={102536},
  year={2021},
  publisher={Elsevier}
}

@article{song2025causal,
  title={Causal keyword driven reliable text classification with large language model feedback},
  author={Song, Rui and Li, Yingji and Tian, Mingjie and Wang, Hanwen and Giunchiglia, Fausto and Xu, Hao},
  journal={Information Processing \& Management},
  volume={62},
  number={2},
  pages={103964},
  year={2025},
  publisher={Elsevier}
}

@article{mutlu2020candidate,
  title={Candidate sentence selection for extractive text summarization},
  author={Mutlu, Begum and Sezer, Ebru A and Akcayol, M Ali},
  journal={Information Processing \& Management},
  volume={57},
  number={6},
  pages={102359},
  year={2020},
  publisher={Elsevier}
}

@inproceedings{lin2004rouge,
  title={Rouge: A package for automatic evaluation of summaries},
  author={Lin, Chin-Yew},
  booktitle={Text summarization branches out},
  pages={74--81},
  year={2004}
}

@inproceedings{10.3115/1073083.1073135,
  title={Bleu: a method for automatic evaluation of machine translation},
  author={Papineni, Kishore and Roukos, Salim and Ward, Todd and Zhu, Wei-Jing},
  booktitle={Proceedings of the 40th annual meeting of the Association for Computational Linguistics},
  pages={311--318},
  year={2002}
}

@inproceedings{zhangbertscore,
  title={BERTScore: Evaluating Text Generation with BERT},
  author={Tianyi Zhang* and Varsha Kishore* and Felix Wu* and Kilian Q. Weinberger and Yoav Artzi},
  booktitle={International Conference on Learning Representations},
  year={2020}
}

@inproceedings{zhao-etal-2019-moverscore,
  title={MoverScore: Text generation evaluating with contextualized embeddings and earth mover distance},
  author={Zhao, Wei and Peyrard, Maxime and Liu, Fei and Gao, Yang and Meyer, Christian M and Eger, Steffen},
  booktitle={Proceedings of the 2019 conference on empirical methods in natural language processing and the 9th international joint conference on natural language processing (EMNLP-IJCNLP)},
  pages={563--578},
  year={2019}
}

@article{laban2022summac,
  title={SummaC: Re-visiting NLI-based models for inconsistency detection in summarization},
  author={Laban, Philippe and Schnabel, Tobias and Bennett, Paul N and Hearst, Marti A},
  journal={Transactions of the Association for Computational Linguistics},
  volume={10},
  pages={163--177},
  year={2022},
  publisher={MIT Press One Rogers Street, Cambridge, MA 02142-1209, USA journals-info~…}
}

@inproceedings{sun-etal-2022-bertscore,
  title={BERTScore is unfair: On social bias in language model-based metrics for text generation},
  author={Sun, Tianxiang and He, Junliang and Qiu, Xipeng and Huang, Xuan-Jing},
  booktitle={Proceedings of the 2022 conference on empirical methods in natural language processing},
  pages={3726--3739},
  year={2022}
}

@inproceedings{bhandari-etal-2020-evaluating,
  title={Re-evaluating evaluation in text summarization},
  author={Bhandari, Manik and Gour, Pranav Narayan and Ashfaq, Atabak and Liu, Pengfei and Neubig, Graham},
  booktitle={Proceedings of the 2020 Conference on Empirical Methods in Natural Language Processing (EMNLP)},
  pages={9347--9359},
  year={2020}
}

@inproceedings{koh-etal-2022-far,
  title={How far are we from robust long abstractive summarization?},
  author={Koh, Huan Yee and Ju, Jiaxin and Zhang, He and Liu, Ming and Pan, Shirui},
  booktitle={Proceedings of the 2022 Conference on Empirical Methods in Natural Language Processing},
  pages={2682--2698},
  year={2022}
}

@inproceedings{mahmoudi-2023-exploring,
  title={Exploring prompting large language models as explainable metrics},
  author={Mahmoudi, Ghazaleh},
  booktitle={Proceedings of the 4th Workshop on Evaluation and Comparison of NLP Systems},
  pages={219--227},
  year={2023}
}

@inproceedings{
paulus2018a,
title={A Deep Reinforced Model for Abstractive Summarization},
author={Romain Paulus and Caiming Xiong and Richard Socher},
booktitle={International Conference on Learning Representations},
year={2018}
}

@inproceedings{10.5555/3304222.3304389,
  title={A reinforced topic-aware convolutional sequence-to-sequence model for abstractive text summarization},
  author={Wang, Li and Yao, Junlin and Tao, Yunzhe and Zhong, Li and Liu, Wei and Du, Qiang},
  booktitle={Proceedings of the 27th International Joint Conference on Artificial Intelligence},
  pages={4453--4460},
  year={2018}
}

@article{stanczak2025societal,
  title={Societal alignment frameworks can improve llm alignment},
  author={Sta{\'n}czak, Karolina and Meade, Nicholas and Bhatia, Mehar and Zhou, Hattie and B{\"o}ttinger, Konstantin and Barnes, Jeremy and Stanley, Jason and Montgomery, Jessica and Zemel, Richard and Papernot, Nicolas and others},
  journal={arXiv preprint arXiv:2503.00069},
  year={2025}
}

@inproceedings{song-etal-2025-learning,
  title={Learning to Summarize from LLM-generated Feedback},
  author={Song, Hwanjun and Yun, Taewon and Lee, Yuho and Oh, Jihwan and Lee, Gihun and Cai, Jason and Su, Hang},
  booktitle={Proceedings of the 2025 Conference of the Nations of the Americas Chapter of the Association for Computational Linguistics: Human Language Technologies (Volume 1: Long Papers)},
  pages={835--857},
  year={2025}
}

@article{brown2020lan,
  title={Language models are few-shot learners},
  author={Brown, Tom and Mann, Benjamin and Ryder, Nick and Subbiah, Melanie and Kaplan, Jared D and Dhariwal, Prafulla and Neelakantan, Arvind and Shyam, Pranav and Sastry, Girish and Askell, Amanda and others},
  journal={Advances in neural information processing systems},
  volume={33},
  pages={1877--1901},
  year={2020}
}

@article{ouyang2022training,
  title={Training language models to follow instructions with human feedback},
  author={Ouyang, Long and Wu, Jeffrey and Jiang, Xu and Almeida, Diogo and Wainwright, Carroll and Mishkin, Pamela and Zhang, Chong and Agarwal, Sandhini and Slama, Katarina and Ray, Alex and others},
  journal={Advances in neural information processing systems},
  volume={35},
  pages={27730--27744},
  year={2022}
}

@inproceedings{kocmi-federmann-2023-large,
  title={Large language models are state-of-the-art evaluators of translation quality},
  author={Kocmi, Tom and Federmann, Christian},
  booktitle={Proceedings of the 24th Annual Conference of the European Association for Machine Translation},
  pages={193--203},
  year={2023}
}

@article{10.1145/1233912.1233913,
  title={The pyramid method: Incorporating human content selection variation in summarization evaluation},
  author={Nenkova, Ani and Passonneau, Rebecca and McKeown, Kathleen},
  journal={ACM Transactions on Speech and Language Processing (TSLP)},
  volume={4},
  number={2},
  pages={4--es},
  year={2007},
  publisher={ACM New York, NY, USA}
}

@inproceedings{wang-etal,
  title={Large language models are not fair evaluators},
  author={Wang, Peiyi and Li, Lei and Chen, Liang and Cai, Zefan and Zhu, Dawei and Lin, Binghuai and Cao, Yunbo and Kong, Lingpeng and Liu, Qi and Liu, Tianyu and others},
  booktitle={Proceedings of the 62nd Annual Meeting of the Association for Computational Linguistics (Volume 1: Long Papers)},
  pages={9440--9450},
  year={2024}
}

@inproceedings{du2023improving,
  title={Improving factuality and reasoning in language models through multiagent debate},
  author={Du, Yilun and Li, Shuang and Torralba, Antonio and Tenenbaum, Joshua B and Mordatch, Igor},
  booktitle={Forty-first International Conference on Machine Learning},
  year={2023}
}

@article{bai2022constitutional,
  title={Constitutional ai: Harmlessness from ai feedback},
  author={Bai, Yuntao and Kadavath, Saurav and Kundu, Sandipan and Askell, Amanda and Kernion, Jackson and Jones, Andy and Chen, Anna and Goldie, Anna and Mirhoseini, Azalia and McKinnon, Cameron and others},
  journal={arXiv preprint arXiv:2212.08073},
  year={2022}
}

@article{NEURIPS2023_91edff07,
  title={Self-refine: Iterative refinement with self-feedback},
  author={Madaan, Aman and Tandon, Niket and Gupta, Prakhar and Hallinan, Skyler and Gao, Luyu and Wiegreffe, Sarah and Alon, Uri and Dziri, Nouha and Prabhumoye, Shrimai and Yang, Yiming and others},
  journal={Advances in neural information processing systems},
  volume={36},
  pages={46534--46594},
  year={2023}
}

@inproceedings{welleck2023generating,
title={Generating Sequences by Learning to Self-Correct},
author={Sean Welleck and Ximing Lu and Peter West and Faeze Brahman and Tianxiao Shen and Daniel Khashabi and Yejin Choi},
booktitle={The Eleventh International Conference on Learning Representations },
year={2023}
}

@inproceedings{maynez-etal,
  title={On faithfulness and factuality in abstractive summarization},
  author={Maynez, Joshua and Narayan, Shashi and Bohnet, Bernd and McDonald, Ryan},
  booktitle={Proceedings of the 58th annual meeting of the association for computational linguistics},
  pages={1906--1919},
  year={2020}
}

@inproceedings{pang-etal-2023-long,
  title={Long document summarization with top-down and bottom-up inference},
  author={Pang, Bo and Nijkamp, Erik and Kry{\'s}ci{\'n}ski, Wojciech and Savarese, Silvio and Zhou, Yingbo and Xiong, Caiming},
  booktitle={Findings of the Association for Computational Linguistics: EACL 2023},
  pages={1267--1284},
  year={2023}
}

@inproceedings{mao-etal-2022-dyle,
  title={DYLE: Dynamic latent extraction for abstractive long-input summarization},
  author={Mao, Ziming and Wu, Chen Henry and Ni, Ansong and Zhang, Yusen and Zhang, Rui and Yu, Tao and Deb, Budhaditya and Zhu, Chenguang and Awadallah, Ahmed and Radev, Dragomir},
  booktitle={Proceedings of the 60th Annual Meeting of the Association for Computational Linguistics (Volume 1: Long Papers)},
  pages={1687--1698},
  year={2022}
}

@article{10.5555/3495724.3495977,
  title={Learning to summarize with human feedback},
  author={Stiennon, Nisan and Ouyang, Long and Wu, Jeffrey and Ziegler, Daniel and Lowe, Ryan and Voss, Chelsea and Radford, Alec and Amodei, Dario and Christiano, Paul F},
  journal={Advances in neural information processing systems},
  volume={33},
  pages={3008--3021},
  year={2020}
}

@inproceedings{wang-etal-2020-asking,
  title={Asking and answering questions to evaluate the factual consistency of summaries},
  author={Wang, Alex and Cho, Kyunghyun and Lewis, Mike},
  booktitle={Proceedings of the 58th annual meeting of the association for computational linguistics},
  pages={5008--5020},
  year={2020}
}

@inproceedings{ding2024evaluation,
  title={Evaluation of question-answering based text summarization using LLM invited paper},
  author={Ding, Junhua and Nguyen, Huyen and Chen, Haihua},
  booktitle={2024 IEEE International Conference on Artificial Intelligence Testing (AITest)},
  pages={142--149},
  year={2024},
  organization={IEEE}
}

@inproceedings{ding2023quality,
  title={Quality evaluation of summarization models for patent documents},
  author={Ding, Junhua and Chen, Haihua and Kolapudi, Sai and Pobbathi, Lavanya and Nguyen, Huyen},
  booktitle={2023 IEEE 23rd International Conference on Software Quality, Reliability, and Security (QRS)},
  pages={250--259},
  year={2023},
  organization={IEEE}
}

@inproceedings{grattafiori2024llama3,
  title={The Llama 3 herd of models},
  author={Grattafiori, Aaron and Dubey, Abhimanyu and Jauhri, Abhinav and Pandey, Abhinav and Kadian, Abhishek and Al-Dahle, Ahmad and Letman, Aiesha and Mathur, Akhil and Schelten, Alan and Vaughan, Alex and others},
  booktitle={Neural Information Processing Systems},
  year={2024},
  organization={Curran Associates}
}

@inproceedings{dang2005overview,
  title={Overview of DUC 2005},
  author={Dang, Hoa Trang},
  booktitle={Proceedings of the document understanding conference},
  volume={2005},
  pages={1--12},
  year={2005}
}

@article{nguyen2024comparative,
  title={A comparative study of quality evaluation methods for text summarization},
  author={Nguyen, Huyen and Chen, Haihua and Pobbathi, Lavanya and Ding, Junhua},
  journal={arXiv preprint arXiv:2407.00747},
  year={2024}
}

@inproceedings{banerjee-lavie-2005-meteor,
  title={METEOR: An automatic metric for MT evaluation with improved correlation with human judgments},
  author={Banerjee, Satanjeev and Lavie, Alon},
  booktitle={Proceedings of the acl workshop on intrinsic and extrinsic evaluation measures for machine translation and/or summarization},
  pages={65--72},
  year={2005}
}

@inproceedings{popovic-2015-chrf,
  title={chrF: character n-gram F-score for automatic MT evaluation},
  author={Popovi{\'c}, Maja},
  booktitle={Proceedings of the tenth workshop on statistical machine translation},
  pages={392--395},
  year={2015}
}

@inproceedings{scialom-etal-2019-answers,
  title={Answers unite! unsupervised metrics for reinforced summarization models},
  author={Scialom, Thomas and Lamprier, Sylvain and Piwowarski, Benjamin and Staiano, Jacopo},
  booktitle={Proceedings of the 2019 Conference on Empirical Methods in Natural Language Processing and the 9th International Joint Conference on Natural Language Processing (EMNLP-IJCNLP)},
  pages={3246--3256},
  year={2019}
}

@article{deutsch-etal-2021-towards,
  title={Towards question-answering as an automatic metric for evaluating the content quality of a summary},
  author={Deutsch, Daniel and Bedrax-Weiss, Tania and Roth, Dan},
  journal={Transactions of the Association for Computational Linguistics},
  volume={9},
  pages={774--789},
  year={2021},
  publisher={MIT Press One Rogers Street, Cambridge, MA 02142-1209, USA journals-info~…}
}

@inproceedings{scialom-etal-2021-questeval,
  title={QuestEval: Summarization asks for fact-based evaluation},
  author={Scialom, Thomas and Dray, Paul-Alexis and Lamprier, Sylvain and Piwowarski, Benjamin and Staiano, Jacopo and Wang, Alex and Gallinari, Patrick},
  booktitle={Proceedings of the 2021 conference on empirical methods in natural language processing},
  pages={6594--6604},
  year={2021}
}

@inproceedings{vasilyev-etal-2020-fill,
  title={Fill in the BLANC: Human-free quality estimation of document summaries},
  author={Vasilyev, Oleg and Dharnidharka, Vedant and Bohannon, John},
  booktitle={Proceedings of the First Workshop on Evaluation and Comparison of NLP Systems},
  pages={11--20},
  year={2020}
}

@article{yang2024qwen2,
  title={Qwen2 Technical Report},
  author={Yang, An and Yang, Baosong and Hui, Binyuan and Zheng, Bo and Yu, Bowen and Zhou, Chang and Li, Chengpeng and Li, Chengyuan and Liu, Dayiheng and Huang, Fei and others},
  journal={eprint arXiv: 2407.10671},
  year={2024}
}

@article{abdin2024phi4,
  title={Phi-4 technical report},
  author={Abdin, Marah and Aneja, Jyoti and Behl, Harkirat and Bubeck, S{\'e}bastien and Eldan, Ronen and Gunasekar, Suriya and Harrison, Michael and Hewett, Russell J and Javaheripi, Mojan and Kauffmann, Piero and others},
  journal={arXiv preprint arXiv:2412.08905},
  year={2024}
}

@article{10.5555/3540261.3542349,
  title={Bartscore: Evaluating generated text as text generation},
  author={Yuan, Weizhe and Neubig, Graham and Liu, Pengfei},
  journal={Advances in neural information processing systems},
  volume={34},
  pages={27263--27277},
  year={2021}
}

@inproceedings{wilber-etal-2021-point,
  title={To point or not to point: Understanding how abstractive summarizers paraphrase text},
  author={Wilber, Matt and Timkey, William and Van Schijndel, Marten},
  booktitle={Findings of the Association for Computational Linguistics: ACL-IJCNLP 2021},
  pages={3362--3376},
  year={2021}
}

@inproceedings{kornilova,
  title={BillSum: A corpus for automatic summarization of US legislation},
  author={Kornilova, Anastassia and Eidelman, Vladimir},
  booktitle={Proceedings of the 2nd Workshop on New Frontiers in Summarization},
  pages={48--56},
  year={2019}
}

@inproceedings{chan2024chateval,
title={ChatEval: Towards Better {LLM}-based Evaluators through Multi-Agent Debate},
author={Chi-Min Chan and Weize Chen and Yusheng Su and Jianxuan Yu and Wei Xue and Shanghang Zhang and Jie Fu and Zhiyuan Liu},
booktitle={The Twelfth International Conference on Learning Representations},
year={2024}
}

@inproceedings{wu2025on,
title={On the Emergence of Position Bias in Transformers},
author={Xinyi Wu and Yifei Wang and Stefanie Jegelka and Ali Jadbabaie},
booktitle={Forty-second International Conference on Machine Learning},
year={2025}
}

@inproceedings{ji-etal-2023-towards,
  title={Towards mitigating LLM hallucination via self reflection},
  author={Ji, Ziwei and Yu, Tiezheng and Xu, Yan and Lee, Nayeon and Ishii, Etsuko and Fung, Pascale},
  booktitle={Findings of the Association for Computational Linguistics: EMNLP 2023},
  pages={1827--1843},
  year={2023}
}

@inproceedings{sharma-etal-2019-bigpatent,
  title={BIGPATENT: A large-scale dataset for abstractive and coherent summarization},
  author={Sharma, Eva and Li, Chen and Wang, Lu},
  booktitle={Proceedings of the 57th annual meeting of the association for computational linguistics},
  pages={2204--2213},
  year={2019}
}

@article{liang2023holistic,
title={Holistic Evaluation of Language Models},
author={Liang, Percy and Bommasani, Rishi and Lee, Tony and Tsipras, Dimitris and Soylu, Dilara and Yasunaga, Michihiro and Zhang, Yian and Narayanan, Deepak and Wu, Yuhuai and Kumar, Ananya and others},
journal={Transactions on Machine Learning Research},
issn={2835-8856},
year={2023}
}

@inproceedings{kim2024prometheus,
title={Prometheus: Inducing Fine-Grained Evaluation Capability in Language Models},
author={Seungone Kim and Jamin Shin and Yejin Cho and Joel Jang and Shayne Longpre and Hwaran Lee and Sangdoo Yun and Seongjin Shin and Sungdong Kim and James Thorne and Minjoon Seo},
booktitle={The Twelfth International Conference on Learning Representations},
year={2024}
}

@inproceedings{el-haj-etal-2020-financial,
  title={The financial narrative summarisation shared task (FNS 2020)},
  author={El-Haj, Mahmoud and Litvak, Marina and Pittaras, Nikiforos and Giannakopoulos, George and others},
  booktitle={Proceedings of the 1st Joint Workshop on Financial Narrative Processing and MultiLing Financial Summarisation},
  pages={1--12},
  year={2020}
}

@article{altmami2022automatic,
  title={Automatic summarization of scientific articles: A survey},
  author={Altmami, Nouf Ibrahim and Menai, Mohamed El Bachir},
  journal={Journal of King Saud University-Computer and Information Sciences},
  volume={34},
  number={4},
  pages={1011--1028},
  year={2022},
  publisher={Elsevier}
}

@article{akter2025comprehensive,
  title={A comprehensive survey on legal summarization: Challenges and future directions},
  author={Akter, Mousumi and {\c{C}}ano, Erion and Weber, Erik and Dobler, Dennis and Habernal, Ivan},
  journal={ACM Computing Surveys},
  volume={58},
  number={7},
  pages={1--32},
  year={2025},
  publisher={ACM New York, NY}
}

\end{document}